\documentclass[sigconf]{acmart}

\AtBeginDocument{%
  }

\copyrightyear{2026}
\acmYear{2026}
\setcopyright{cc}
\setcctype{by}
\acmConference[CHI '26]{Proceedings of the 2026 CHI Conference on Human Factors in Computing Systems}{April 13--17, 2026}{Barcelona, Spain}
\acmBooktitle{Proceedings of the 2026 CHI Conference on Human Factors in Computing Systems (CHI '26), April 13--17, 2026, Barcelona, Spain}
\acmPrice{}
\acmDOI{10.1145/3772318.3790770}
\acmISBN{979-8-4007-2278-3/2026/04}

\usepackage[utf8]{inputenc}
\usepackage{enumitem}
\usepackage{makecell}
\usepackage{multirow}
\usepackage{subcaption}
\usepackage{rotating}

\definecolor{bgcolor}{RGB}{230, 240, 255}

\newif\ifdraft
\draftfalse

\ifdraft
  \newcommand{\todo}[1]{\textsf{\textbf{\textcolor{red}{[TODO: #1]}}}}
\else
  \newcommand{\todo}[1]{}
\fi

\ifdraft
  \newcommand{\rev}[1]{\textcolor{blue}{#1}}
\else
  \newcommand{\rev}[1]{#1}
\fi

\begin{document}

\title{"I think this is fair": Uncovering the Complexities of Stakeholder Decision-Making in AI Fairness Assessment}

\author{Lin Luo}
\authornote{Corresponding Author}
\orcid{https://orcid.org/0000-0002-0310-3158}
\affiliation{%
  \department{School of Computing Science}
  \institution{University of Glasgow}
  \city{Glasgow}
  \country{United Kingdom}
}
\email{l.luo.1@research.gla.ac.uk}

\author{Yuri Nakao}
\orcid{https://orcid.org/0000-0002-6813-9952}
\affiliation{%
  \department{Fujitsu Research}
  \institution{Fujitsu Limited}
  \city{Kawasaki}
  \country{Japan}
}
\email{nakao.yuri@fujitsu.com}

\author{Mathieu Chollet}
\orcid{https://orcid.org/0000-0001-9858-6844}
\affiliation{%
  \department{School of Computing Science}
  \institution{University of Glasgow}
  \city{Glasgow}
  \country{United Kingdom}
}
\email{mathieu.chollet@glasgow.ac.uk}

\author{Hiroya Inakoshi}
\orcid{https://orcid.org/0000-0003-4405-8952}
\affiliation{%
  \department{Fujitsu Research}
  \institution{Fujitsu Limited}
  \city{Kawasaki}
  \country{Japan}
}
\email{inakoshi.hiroya@fujitsu.com}

\author{Simone Stumpf}
\orcid{https://orcid.org/0000-0001-6482-1973}
\affiliation{%
  \department{School of Computing Science}
  \institution{University of Glasgow}
  \city{Glasgow}
  \country{United Kingdom}
}
\email{simone.stumpf@glasgow.ac.uk}

\renewcommand{\shortauthors}{Luo et al.}

\begin{abstract}
Assessing fairness in artificial intelligence (AI) typically involves AI experts who select protected features, fairness metrics, and set fairness thresholds to assess outcome fairness. However, little is known about how stakeholders, particularly those affected by AI outcomes but lacking AI expertise, assess fairness. To address this gap, we conducted a qualitative study with 26 stakeholders without AI expertise, representing potential decision subjects in a credit rating scenario, to examine how they assess fairness when placed in the role of deciding on features with priority, metrics, and thresholds. We reveal that stakeholders' fairness decisions are more complex than typical AI expert practices: they considered features far beyond legally protected features, tailored metrics for specific contexts, set diverse yet stricter fairness thresholds, and even preferred designing customized fairness. Our results extend the understanding of how stakeholders can meaningfully contribute to AI fairness governance and mitigation, underscoring the importance of incorporating stakeholders' nuanced fairness judgments.
 
\end{abstract}



\maketitle

\section{Introduction}
\label{sec: introduction}
Fairness in artificial intelligence (AI) has gained significant attention due to potential harms caused by relying on predicted outcomes, such as loan applications \cite{10.1145/3490354.3494408} and university admission \cite{10.1145/3351095.3372867}. AI fairness has been conceptualized along three main perspectives: \textit{interactional fairness}, i.e., focusing on fair interactions with individuals \cite{bies1987interactional}; \textit{procedural fairness}, i.e., fair processes leading to an AI system's outcomes \cite{Robert2020Designing,10.1145/3359284}; and \textit{outcome fairness}, i.e., evaluating whether the outcomes of an AI system are fair \cite{rawls2001justice}. Of these, outcome fairness is the most commonly assessed \cite{Robert2020Designing, 10.5555/3504035.3504042}, and it is also the perspective we seek to explore in this work.

In practice, AI outcome fairness has often been measured quantitatively through fairness metrics that evaluate whether a model treats individuals or groups in a fair manner \cite{10.1145/3194770.3194776}. Typically, AI fairness researchers and practitioners with AI-related expertise whose task it is to design fair AI systems or mitigate existing AI systems so they are fair \cite{FairnessAssessmentPractionerPerspective}--who we refer to in this work as AI experts--identify features that they would like to assess and protect, and then choose fairness metrics alongside thresholds that determine when outcomes are considered fair \cite{hort2023bias,10.1145/2783258.2783311, Saplicki2022}. While this common expert-driven practice simplifies processes and achieves nominal technical AI fairness, it rarely aligns with human-centered fairness, which goes beyond AI expert perspectives and requires AI systems to reflect the values and priorities that diverse individuals and groups see as essential to fairness \cite{10.1145/3025453.3025884, earnfairness}. This misalignment arises because people interpret fairness differently \cite{landers2023auditing}. Recent studies further show that stakeholders without AI expertise, that is those without formal training or professional experience in AI model development, often differ from AI experts in defining AI fairness \cite{earnfairness, doi:10.1080/10447318.2022.2067936}. Moreover, even AI experts themselves report persistent challenges in determining which features to protect and which fairness metrics to apply \cite{FairnessAssessmentPractionerPerspective}. 

Bridging the gap between technical fairness and human-centered fairness is now becoming an increasingly urgent agenda for Human–Computer Interaction (HCI) and AI research \cite{10.1145/3711079, 10.1145/3025453.3025884, earnfairness, 10.1145/3630106.3659044, 10.1145/3544548.3581378}. Although existing research has explored how AI experts assess fairness \cite{10.1145/3544548.3581227} and the involvement of domain experts \cite{doi:10.1080/10447318.2022.2067936}, psychologists and ethics watchdogs \cite{landers2023auditing}, the voices of decision subjects, those directly affected by AI outcomes who often lack AI expertise, remain strikingly absent. As AI permeates everyday domains, everyone in society increasingly becomes decision subjects, yet they still have almost no effective avenues to participate in fairness assessments \cite{landers2023auditing, doi:10.1080/10447318.2022.2067936}. By overlooking these crucial voices, AI fairness may fall short of public expectations and lead to failures in real-world deployment \cite{2024FaccTAIFailureCards,10.1145/3334480.3375158, book/EthicalAlgorithm/10.5555/3379082, 10.1145/3173574.3174014}. 

We argue that these stakeholders on equal footing with AI experts, need to be able to engage directly with questions of \textit{whom to protect} (features), \textit{how to protect} (metrics), and \textit{to what extent} (thresholds). Such involvement not only increases transparency and fosters public trust but also contributes to actionable fairness strategies, provides crucial guidance to AI experts, and helps avoid the additional biases that may arise when stakeholders' preferences are mediated.

Our work pursues this aim by supporting and uncovering how those who are affected by AI outcomes and also lack AI expertise (whom we refer to as lay stakeholders, or simply stakeholders in this work) assess AI outcome fairness. However, there remain four key research gaps that need to be addressed. First, current HCI and fairness practice led by AI experts may result in overly narrow definitions of fairness and introduce biases stemming from experts' choices of features and metrics \cite{10.1145/3411764.3445308}. Indeed, to which features stakeholders pay attention has not been sufficiently investigated, even though previous research indicates that stakeholders have a broader fairness concerns, including features such as \textit{income} in loan scenarios \cite{10.1145/3514258}. Second, current research does not capture how these features are prioritized by stakeholders to understand their fairness trade-offs. Third, current studies have attempted to solicit metric preferences from stakeholders \cite{FaccT2024MetricPreferenceInvestigation, wong2020democratizing,10.1145/3411764.3445308,Saxena/10.1145/3306618.3314248,2024crowdsource/10.1145/3640543.3645209,ilvento:LIPIcs.FORC.2020.2,Yokota2022Toward} but primarily focus on soliciting stakeholders' most preferred fairness metric and overlook how stakeholders perceive fairness nuances at the level of specific features \cite{10.1145/3411764.3445308, 2024crowdsource/10.1145/3640543.3645209,MathematicalNotions/10.1145/3292500.3330664, Saxena/10.1145/3306618.3314248}. This limits opportunities for stakeholders to articulate their specific fairness needs \cite{doi:10.1080/10447318.2022.2067936}, and fails to uncover the complexity of how metric preferences interact with different features \cite{2024crowdsource/10.1145/3640543.3645209, 10.1145/3411764.3445308}. Last, only limited research has engaged stakeholders in deciding acceptable fairness thresholds \cite{earnfairness}, but those established by legal standards or determined by AI experts may again not necessarily align with stakeholder expectations. Therefore, general studies that begin with and focus solely on metric preferences \cite{Saxena/10.1145/3306618.3314248,10.1145/3514258, MathematicalNotions/10.1145/3292500.3330664,2024crowdsource/10.1145/3640543.3645209,FaccT2024MetricPreferenceInvestigation, earnfairness, 10.1145/3411764.3445308,10.1145/3357236.3395528, doi:10.1080/10447318.2022.2067936} may offer AI experts some insight into how stakeholders define fairness. However, such an approach may not reveal which aspects of fairness stakeholders believe should be protected and prioritized, or whether the AI system is perceived as fair under the applied metrics, thereby possibly failing to align fairness with stakeholder values. 

To address these gaps, we reframe AI fairness assessment as a participatory process that empowers stakeholders to perform tasks traditionally reserved for AI experts. This process centers on three core lenses: (1) selecting and ranking features to determine \textit{whom to protect}, (2) mapping features to fairness metrics to define \textit{how to protect them}, and (3) setting thresholds to specify \textit{what level of unfairness is acceptable}. We investigated this process through a qualitative study with 26 participants who lacked AI expertise but represented potential decision subjects. Each participant engaged in fairness assessment tasks through an interactive prototype system we designed to support this process. We chose a commonly used credit rating scenario to ground our investigation. By observing stakeholders perform fairness assessment, we show the complexity of their fairness decision-making while generating insights that can directly inform AI mitigation in real-world systems. We address the following research questions: 
\begin{itemize}
    \item \textbf{RQ1} How do stakeholders identify features that are crucial for assessing AI outcome fairness, and how do they prioritize them?
    \item \textbf{RQ2} How do stakeholders make metric choices to best reflect their perceived fairness for different features, and how do they set fairness thresholds? 
\end{itemize}

The contributions of our work are as follows:
\begin{itemize}
    \item We provide empirical evidence showing how stakeholders without AI expertise assess AI outcome fairness across features, metrics, and fairness thresholds in a credit-rating context. Our findings shed light on the complexity of their reasoning and the generic patterns that differ from common expert practices.
    \item We offer new insights into the importance of unprotected features and custom fairness metrics for stakeholders.
    \item We derive actionable design implications for interactive tools and governance processes that make AI fairness assessment more accessible to stakeholders, enable stakeholders to conduct AI fairness assessment alongside experts, and leverage their insights to inform fairness mitigation in AI systems.
\end{itemize}

This paper is structured as follows. Section~\ref{sec: Related Work} reviews existing research and practices measuring AI fairness from an algorithmic perspective and stakeholder-centered approaches to AI fairness, identifying gaps our study addresses. Section~\ref{sec: Methods} details our qualitative study in a credit rating scenario and the prototype system that operationalized outcome fairness assessment for lay stakeholders. Section~\ref{sec: Results} presents the results of each research question, and Section~\ref{sec: Discussion} reflects on stakeholder decision complexity, with design implications for future fairness tools.

\section{Related Work}
\label{sec: Related Work}
This section presents common ways of identifying unfairness and prevalent practices among AI experts to investigate the fairness of AI models. We then provide an overview of human-centered studies of fairness.

\subsection{Measuring AI Fairness}
\label{sec: AI experts assessment pattern}
Among the three perspectives on AI fairness, interactional fairness is relatively hard to measure, since it considers whether people understand the information an AI system provides and feel respected in the interaction \cite{Robert2020Designing}. Procedural fairness has gained attention, though standardized metrics remain limited; prior works often examine which features should be used, whether inappropriate or sensitive features (e.g., race, gender) are included, or ways to restrict their use \cite{10.1145/3461702.3462585, 10.5555/3504035.3504042}. Outcome fairness, in comparison, is the perspective most commonly translated into quantitative assessments and AI experts have proposed 109 different fairness metrics \cite{hort2023bias}, with more than 20 widely used ones \cite{10.1145/3194770.3194776}, addressing different fairness needs to ensure equitable outcomes for individuals or groups \cite{10.1145/3494672,10.1145/3457607,hort2023bias}.

Of major importance in AI outcome fairness are \textit{protected features}, which are safeguarded by anti-discrimination laws \cite{hort2023bias,10.1145/3457607,10.1145/3194770.3194776, rabonato_systematic_2025}, for example, protections against discrimination based on age, gender, and race under the United Kingdom's Equality Act 2010 \cite{EqualityAct2010}, EU Law \cite{EUCharter2012}, or United States' anti-discrimination laws in employment \cite{USLaw}. Some \textit{sensitive features}, such as ``foreign worker'' \cite{10.1145/3457607,4909197,10.1145/3287560.3287589}, while not strictly covered by law, are also sometimes deemed important in certain contexts as they may be related to protected characteristics such as race.

Group fairness metrics aim to measure equal outcomes across demographic groups in these selected protected features. Some common metrics include:

Demographic Parity \cite{NIPS2017_a486cd07} or, Statistical Parity \cite{10.1145/2090236.2090255}, which measure whether protected and unprotected groups have equal probabilities of receiving the favorable outcome from AI;
Equal Opportunity \cite{10.5555/3157382.3157469}, which measures protected and unprotected groups have the same probability of receiving the favorable outcome from AI for individuals who truly qualify for it;
Predictive Equality \cite{10.1145/3097983.3098095}, which measures whether protected and unprotected groups have the same probability of incorrectly receiving the favorable AI outcome for individuals who do not actually qualify for it;
Equalized Odds \cite{10.5555/3157382.3157469}, which requires satisfying both Equal Opportunity and Predictive Equality; 
Outcome Test \cite{articleoutcometest}, which measures whether individuals predicted to receive the favorable AI outcome from both groups have the same probability of truly qualifying for it; and
Conditional Statistical Parity \cite{10.1145/3097983.3098095}, which measures whether protected and unprotected groups have an equal probability of receiving the favorable outcome from AI, conditioned on ``legitimate'' features. Legitimate features are context-relevant attributes that can influence an individual's AI prediction outcomes; for instance, in credit rating scenarios, legitimate features could be factors affecting an applicant's creditworthiness, such as credit amount \cite{10.1145/3194770.3194776}.
     
Individual fairness metrics focus on similar individuals receiving similar outcomes from AI \cite{10.1145/3457607, 10.1145/2090236.2090255}. For example, Counterfactual Fairness \cite{NIPS2017_a486cd07} measures whether an AI outcome remains unchanged when an individual's protected features are altered, when all other feature values remain the same, while Consistency \cite{pmlr-v28-zemel13, 10.1145/2090236.2090255} ensures that similar individuals receive consistent outcomes from the AI.

AI experts usually choose to assess either individual or group fairness based on the application context, legal requirements, and practical feasibility \cite{10.1145/3287560.3287598, conflict/10.1145/3351095.3372864}. However, individual fairness is difficult to implement and scale due to the lack of well-defined and generalizable methods for measuring individual similarities, making it less common in high-stakes scenarios like job hiring or criminal justice where addressing systemic discrimination is critical \cite{conflict/10.1145/3351095.3372864}. The prevailing practice among AI experts is to evaluate group fairness, often by limiting the assessment to a single protected feature like gender \cite{10.1145/3641276}, or extending it to others such as race or age to meet legal and ethical standards. Then, AI experts choose metrics to evaluate these features on. Typically, AI experts tend to favor popular, statistically-based, and easy-to-implement metrics when assessing fairness and mitigating bias \cite{conflict/10.1145/3351095.3372864, hort2023bias}, such as those described earlier or legally defined standards like US law's Demographic Parity Ratio \cite{10.1145/2783258.2783311}. Sometimes, if available, metrics are chosen from a standard metric set prevalent in the AI expert's organization or they are reused from a previously conducted evaluation \cite{racialBiasPractitionerStandardMetrics, FairnessAssessmentPractionerPerspective}. When more than one feature is selected for measuring group fairness, AI experts often adopt a ``one-size-fits-all'' metric approach by applying the same metric or the same set of metrics to these relevant protected features, often across different scenarios and contexts \cite{10.1145/3442188.3445902, 10.1145/3531146.3533225, 10.1145/2783258.2783311, 10.1145/3287560.3287586, 10.1145/3357384.3357974}. Nearly half of existing studies mitigating fairness issues rely on a single metric \cite{hort2023bias}, reflecting limited fairness assessment.

When multiple features and metrics are in play, optimizing one fairness metric may unintentionally deteriorate another metric \cite{Kleinberg2016InherentTI, 10.1145/3494672, hort2023bias}, and improving fairness for one feature might worsen the same metric for another feature \cite{10.1145/2783258.2783311}. AI experts therefore often face fairness trade-offs and must make critical decisions on which features or groups \cite{10.1145/2783258.2783311} and fairness metrics \cite{10.1145/3194770.3194776, 10.1145/3457607, 10.1145/3494672} to prioritize. Such fairness trade-offs often lead AI experts to prioritize either group fairness or individual fairness, but rarely both, due to practical challenges and the conflicts described above \cite{10.1145/3544548.3581227, hort2023bias}. 

In addition to selecting metrics and features, AI experts must also make decisions regarding fairness thresholds \cite{Gupta2021TransitioningFR}. These thresholds determine acceptable levels of unfairness when assessing fairness, often shaped by domain-specific constraints and legal requirements. Current practice typically involves aligning thresholds with legal mandates, such as a minimum demographic parity ratio (DPR) of 80\% \cite{10.1145/2783258.2783311}, or adhering to a widely accepted standard that allows up to a 10\% difference between groups \cite{Gupta2021TransitioningFR}.  

To support these practices of AI experts, various tools have been developed. Notable open-source toolkits include IBM's AI Fairness 360 \cite{8843908}, Google's What-If \cite{8807255}, Microsoft's Fairlearn \cite{weerts2023fairlearn}, DALEX \cite{10.5555/3546258.3546472}, as well as Aequitas \cite{saleiro2018aequitas}. There are also some research prototypes, such as FairSight \cite{8805420}, Silva \cite{10.1145/3313831.3376447}, FairVis \cite{8986948}, and FairRankVis \cite{9552229} that help with the exploration and assessment of AI fairness. However, these tools typically only include commonly-used metrics influenced by the practices of AI experts. More importantly, what is currently absent is the integration of stakeholders without AI expertise, whose perspectives are essential to understanding the societal impact of AI fairness.

\subsection{Stakeholders' Perspectives of AI Fairness}
Models deemed fair from a technical perspective may fail to be perceived as fair in real-world contexts due to their misalignment with societal values and lack of social acceptability \cite{whittaker2018ai,10.1145/3411764.3445308,10.1145/3334480.3375158}. Previous work highlighted that stakeholder input to guide fairness decision-making is crucial \cite{FairnessAssessmentPractionerPerspective, 10.1145/3025453.3025884, 10.1145/3334480.3375158,10.1145/3631700.3664912}. Thus, understanding how stakeholders perceive and assess AI fairness is essential. 

Prior work exploring stakeholder involvement in AI fairness has explored many different areas: their fairness perceptions around procedural fairness and interactional fairness \cite{10.1145/3544548.3581161}; preferred fairness metrics \cite{earnfairness, 10.1145/3411764.3445308}; fairness-accuracy \cite{10.1145/3357236.3395528,Yokota_2025_Towards} or fairness-utility trade-offs \cite{10.1145/3375627.3375862, 10.1145/3357236.3395528,hiranandani_fair_2020}; and the factors that shape how stakeholders perceive fairness in AI systems \cite{10.1145/3313831.3376813}. 

Outcome fairness, particularly fairness metric preference, has emerged as a dominant focus in recent human-centered fairness research. Existing work typically aims to elicit stakeholders' metric preferences over predefined metrics and identify the most favored one across stakeholder groups, such as domain experts \cite{10.1145/3411764.3445308,10.1145/3357236.3395528, doi:10.1080/10447318.2022.2067936}, policymakers \cite{10.1145/3375627.3375862}, and general users or the public \cite{Saxena/10.1145/3306618.3314248,10.1145/3514258, MathematicalNotions/10.1145/3292500.3330664,2024crowdsource/10.1145/3640543.3645209,FaccT2024MetricPreferenceInvestigation, earnfairness}. Meanwhile, there is also growing interest in investigating metric preferences under different contexts, such as child maltreatment predictions \cite{10.1145/3411764.3445308}, loan decisions \cite{earnfairness,10.1145/3514258, doi:10.1080/10447318.2022.2067936,Saxena/10.1145/3306618.3314248}, crime risk prediction \cite{MathematicalNotions/10.1145/3292500.3330664}, healthcare \cite{MathematicalNotions/10.1145/3292500.3330664,2024crowdsource/10.1145/3640543.3645209}, and fair ranking in scholarship distribution \cite{FaccT2024MetricPreferenceInvestigation}. Efforts have also been made to explore whether the exposure of protected information (e.g., race) \cite{Saxena/10.1145/3306618.3314248} or varying risk levels within contexts influences stakeholders' preferences for fairness metrics \cite{2024crowdsource/10.1145/3640543.3645209}. So far, most research investigating how stakeholders could participate in AI fairness assessment by soliciting their fairness preferences has typically avoided confronting people without AI expertise with the mathematical underpinnings of fairness, such as the logic of fairness metrics, model mechanism and parameters  \cite{MathematicalNotions/10.1145/3292500.3330664,2024crowdsource/10.1145/3640543.3645209,10.1145/3631700.3664912,Saxena/10.1145/3306618.3314248}, trying to side-step and simplify the complex decision-making required by AI experts. However, this prevents stakeholders from directly participating in fairness assessments, and may also introduce biases when their preferences are mediated.

Less attention has been placed on investigating what features matter to stakeholders regarding assessing outcome fairness. Current research is primarily focused on demographic groups which are defined by protected features, such as race equality \cite{FaccT2024MetricPreferenceInvestigation,10.1145/3411764.3445308,2024crowdsource/10.1145/3640543.3645209, Saxena/10.1145/3306618.3314248}. Although some studies suggest that non-protected features also play a significant role in AI fairness \cite{earnfairness}, research is lacking on how stakeholders perceive fairness regarding non-protected features and how they are prioritized when multiple features are considered. While numerous studies have focused on fairness-accuracy \cite{10.1145/3357236.3395528} or fairness-utility trade-offs \cite{10.1145/3375627.3375862, 10.1145/3357236.3395528,hiranandani_fair_2020}, inherent fairness trade-offs within features are particularly important yet remain underexplored. 

Further, because fairness is a highly context-dependent concept, only a few studies have explored how stakeholders match different metrics to different features. For example, within a limited set of three group fairness metrics, existing research showed that participants maintained a consistent metric preference across various protected features in a child maltreatment risk prediction scenario \cite{10.1145/3411764.3445308}. However, it remains unclear whether this consistency persists in other high-stakes domains or when stakeholders are presented with a wider range of fairness metrics or opportunities to express new metrics themselves.

Last, stakeholders' acceptable levels of unfairness remained largely unexplored in existing research. Recent research has begun exploring category-level fairness ``thresholds'', where stakeholders set a unified threshold across multiple features and metrics for both individual and group fairness \cite{earnfairness}. However, more fine-grained fairness thresholds remain unexplored.  

In summary, while significant progress has been achieved, there is a need to investigate how to align technical AI fairness with human-centered notions of fairness; in short, we need to better understand how stakeholders perceive and assess AI outcome fairness. 
Building on this foundation, our work empowers stakeholders to perform fairness assessment tasks in ways similar to AI experts, and we examine how stakeholders approach fairness decision-making, uncovering its complexities across feature selection and ranking, feature-metric pairing, and fairness threshold setting. By taking a human-centered perspective, we aim to support algorithmic fairness by incorporating the complex fairness needs of stakeholders.

\section{Methods}
\label{sec: Methods}
We conducted an in-person qualitative user study in which participants were led through a facilitated individual session by a researcher. During the session, they interacted with a prototype system that supported them to choose features, pair features with metrics, and indicate fairness thresholds, ranking their choices to reflect fairness trade-offs. As they worked in the session, they were thinking aloud to expose their decision-making.

\subsection{Scenario Setup: Dataset and AI Model}
Given that our participants were lay stakeholders, we carefully selected a scenario, a dataset, and an AI model that would allow them to meaningfully engage with fairness assessment without imposing excessive cognitive load. To ground the study in a concrete, high-stakes context, we chose a credit rating scenario, a decision-making domain familiar to the public and consequential in real life.

Within this scenario, we employed the German Credit Dataset \cite{dataset}, one of the most frequently used datasets in fairness research \cite{FairnessDatasetSurvey, de2025towards, pagano2023bias}, including HCI studies \cite{10.1145/3613904.3642627, earnfairness} where lay stakeholders need the ability to reflect on fairness in an understandable setting. This dataset presents a relatively simple binary classification task which classifies individuals as having ``Good Credit'' or ``Bad Credit''. It contains 1,000 instances and 20 features, including multiple legally protected and unprotected features, enabling us to observe participants' fairness assessment behavior while avoiding information overload. It includes two legally protected features (`Age' and `Gender' \cite{EUCharter2012}) and 18 unprotected features. `Age' was converted into a binary feature, with individuals under 25 categorized as the protected group through thresholding at 25 (i.e., $\text{age} < 25$ and $\text{age} \geq 25$ \cite{FairnessDatasetSurvey, 4909197, 10.1145/3287560.3287589}). `Gender' was extracted from the `personal status and gender' feature, designating `female' as the protected group \cite{FairnessDatasetSurvey}. Sometimes, `Foreign Worker' is treated as a sensitive feature in the credit context \cite{4909197, 10.1145/3287560.3287589} because it can indirectly reveal legally protected features like race, but we did not specifically call it out in our study, leaving our participants to decide on its significance. The unprotected features were grouped into at most four distinct categories to avoid data sparsity, while numerical features were divided into up to four bins as needed. Appendix~\ref{app:german_credit_features} provides detailed descriptions of each of the 20 features that participants could refer to, based on the original dataset documentation \cite{dataset, FairnessDatasetSurvey}.

We chose logistic regression as it is widely used in credit rating, where complex models add little accuracy \cite{hort2023bias}. Its interpretability also helps participants provide fairness feedback \cite{10.1016/j.inffus.2019.12.012, 10.1145/3411764.3445308, 10.1145/3514258}. We trained the model using 5-fold cross-validation on all 1000 instances and selected one fold of 200 instances as our test dataset for user exploration. On the selected test dataset, we achieved a reasonable level of AI predictive performance with an accuracy of 0.76. 

Meanwhile, the test dataset can allow participants to explore diverse fairness assessment results across different features and metrics, ranging from 0\% group differences (indicating no unfairness between groups) to 100\% group differences (indicating maximum unfairness between groups).

\subsection{Participant Recruitment}
We recruited 30 participants for the study through public social media and university channels. Participants were potential decision subjects, representing individuals who could be impacted by AI-driven decisions in real-world scenarios. To capture the perspectives of those without AI expertise, we specifically included individuals without formal training or professional experience in AI model development (e.g., computer science education, machine learning engineering, or data science roles).  

We conducted the study locally through in-person sessions. While this approach allowed us to closely observe participants' decisions and reasoning, we recognize its regional limitation and therefore adopted purposive sampling, to prioritize diversity in nationality (resulting in 10 different nationalities) and profession (resulting in 11 different types), with educational backgrounds ranging from secondary school (n=1) to bachelor's degree (n=20), master's degree (n=6), and doctoral degree (n=3), considering that different lived experiences and cultural backgrounds might influence people' views on fairness. Then, we tried to balance gender (13 male and 17 female participants) and age. Ages ranged from 21 to 76 (19 aged 20–30 years, 8 aged 30–40 years, 2 aged 40–50 years, and 1 over 50). We note that our participants were relatively young; however, this aligns with evidence from multiple countries showing high loan application activity among younger age groups \cite{LoanUk, LoanUSA, loanAus}. Moreover, our sample balanced loan experience, with 14 participants reporting prior loan applications and 16 without such experience. 

It is important to note that AI fairness can be perceived and conceptualized from different perspectives. Our study focuses on uncovering how participants make outcome fairness assessments, a perspective that aligns with current popular auditing practice by AI experts: \textit{whom to protect} (features),\textit{how to protect them} (metrics), and \textit{to what extent} (thresholds). However, we cannot assume that all participants hold or prefer to assess fairness through this perspective. Indeed, among the 30 participants we recruited, we did observe \rev{differences in preferred fairness perspectives}. We accounted for this in our data analysis (Section~\ref{sec: data collection and analysis}). Full participant information is available in Table~\ref{tab:Participant} in Appendix~\ref{app:Participant Information}.

Every one of the 30 recruited participants received a \pounds 15 Amazon voucher as compensation for their time and effort. We carefully considered the ethical implications of our study, particularly focusing on obtaining informed consent, privacy, and confidentiality. This study was reviewed and approved by the University of Glasgow College of Science \& Engineering Ethics Committee (Application Number: 300230183) as a low-risk study.

\subsection{Procedure}
\label{sec:Procedure}
\subsubsection{Preparation Phase}
Each participant completed the study individually in an in-person session with a researcher. Each participant first went through a 15-minute Preparation Phase where they reviewed the participant information sheet, signed a consent form, and completed a background questionnaire to provide demographic information. 

Prior work has shown that people tend to view an AI system as more fair when they personally receive a favorable outcome \cite{10.1145/3313831.3376813}, for example, a ``Good Credit'' decision in our credit rating scenario. To ensure an objective assessment and remain unaffected by favorable or unfavorable AI outcomes, our participants, as decision subjects, were not \rev{told that any specific decision label applied to themselves (i.e., whether they would personally receive ``Good Credit'' or ``Bad Credit'') when assessing fairness}.

\subsubsection{Main Phase: Three Tasks for Assessing Fairness}
After this setup, each participant proceeded individually through three tasks in the Main Phase, using our prototype system (see Section~\ref{sec:prototype system}). We established the timing for each task through pilot studies, which confirmed that participants had sufficient time to complete the tasks and reflect on fairness. Before each task, the researcher described what the task involved, introduced the relevant prototype functionalities, and played a 2–3 minute video showing how to use the prototype for that task. For example, in the feature selection task, the video showed how to use the prototype to select and drag features for ranking.

\textbf{Task 1: Feature Selection \& Ranking} (average 20 minutes) - Participants were asked, ``\textit{Which features do you consider crucial for assessing AI fairness? Please select the feature first, then drag to rank.}'' In response to this question, participants identified whom to protect, that is, which features they believed should be protected in fairness evaluations, and ordered these features according to the priority they placed on fairness across them. 

\textbf{Task 2: Mapping Features to Metrics with Thresholds} (average 55 minutes) - For features they had identified in Task 1, participants were asked, ``\textit{How would you like to assess the fairness of each feature? You can select from the existing fairness definitions or create your own custom fairness definition.}'' Participants first decided how to protect them by determining the fairness metric they considered most appropriate for each feature. When participants defined a custom metric, the researcher carefully documented their description in real time to ensure the definition was captured accurately. Participants were then asked to set a fairness threshold for each determined feature–metric pair. In our study, fairness thresholds were defined as the maximum allowable deviation from the optimal fairness outcome of the current metric, ranging from 0\% to 100\%. We explained it to them in very simple terms, ``\textit{What is the maximum level of unfairness you are willing to accept? You can think of your maximum unfairness tolerance as a range from 0\% to 100\%. It goes from 0\% to 100\%. If you set 0\%, that means you expect a perfectly fair result on this feature-metric pair, and you won't accept any unfairness at all. The higher the percentage you set, the more unfairness you are saying you can live with. And based on what you set, if the system's evaluation ever goes beyond your threshold, that means, by your standard, the system is unfair.}'' 

\textbf{Task 3: Fairness Re-Ranking} (average 10 minutes) - In the final task, participants were asked to reflect on whether their original priorities still held after assigning fairness metrics. To guide this step, the researcher asked, ``\textit{Since fairness metrics might conflict, it is impractical to satisfy all of them simultaneously. After exploring both features and fairness metrics, you can now re-rank your Feature-Metric pairs based on your fairness priority, or retain your original ranking (i.e., feature ranking).}''

During each task, participants were free to ask questions at any time, and the researcher provided clarifications when needed to ensure that task instructions and prototype usage were fully understood. The researcher's role was strictly limited to clarifying procedures and prototype functionality; no guidance or opinions about fairness itself were introduced, ensuring that participants' assessment reflected their own perspectives. 

Participants were asked to \textit{think aloud} while making decisions, allowing their reasoning to be captured in real time. When appropriate, the researcher posed follow-up questions that prompted participants to elaborate, helping to capture richer explanations and increasing confidence in the findings.

\subsubsection{Post-Task Phase}
After the Main Phase, participants completed a post-survey questionnaire, including ten questions on the System Usability Score (SUS) \cite{SUSEvaluation} of the prototype system, and four open-ended questions to collect detailed feedback on participants' experiences (Appendix~\ref{app:Post Questionnaires}). These are used to supplement our results (Section~\ref{sec: Results}). On average, participants spent nearly two hours ($\approx$110 minutes) completing the study.

\begin{figure*}[htbp]
\centering
\includegraphics[width=0.95\linewidth]{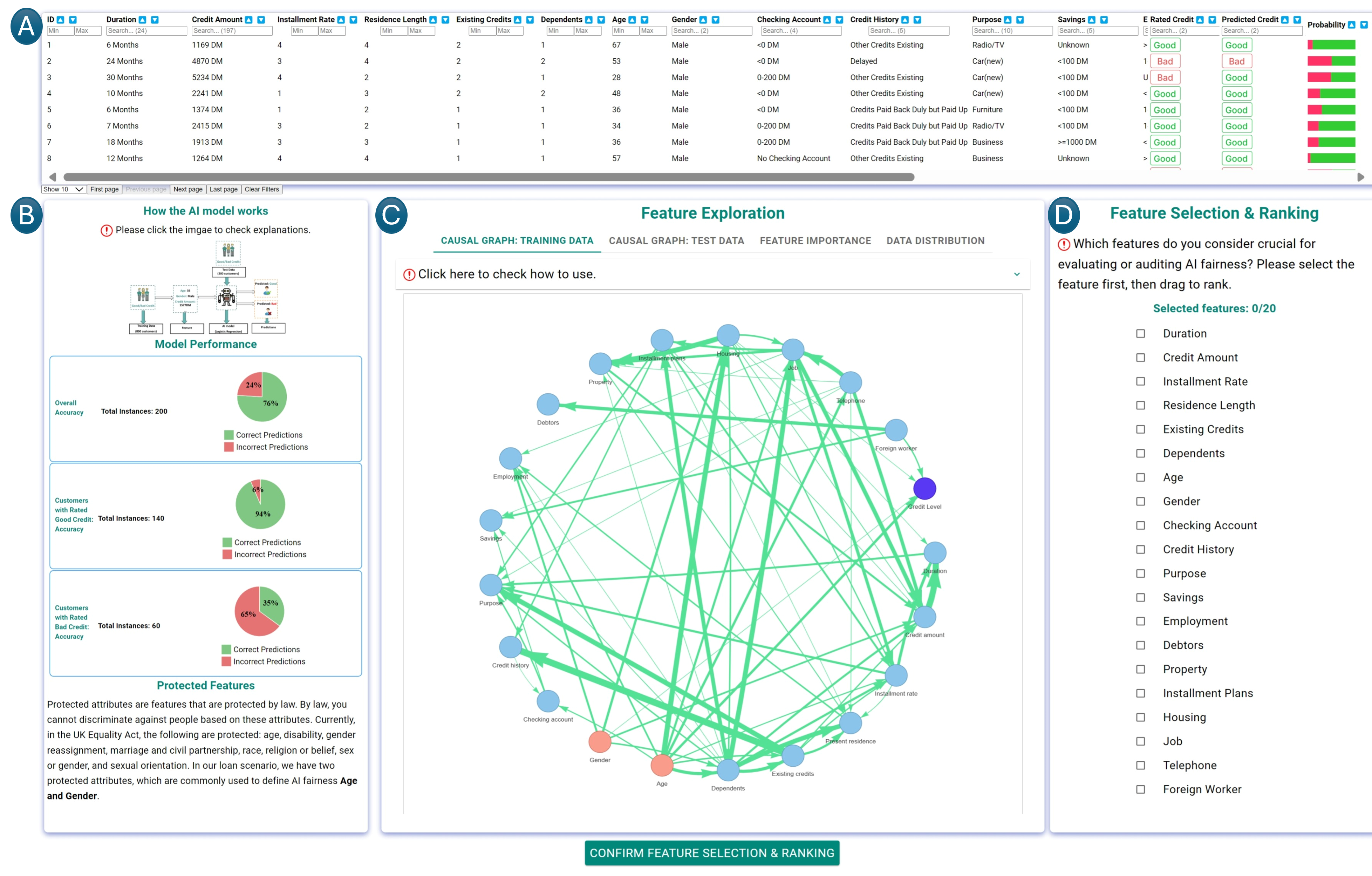}
\caption{\textbf{Prototype System: Dashboard (also, Task 1).}
\textbf{Component A} presents the test dataset in a tabular format, \rev{where each row represents one case (i.e., applicant) of the test dataset, displaying each applicant's feature values, the ground-truth label, and the AI's predicted outcome}.
\textbf{Component B} provides AI model explanations, performance, and information about protected features.
\textbf{Component C} contains four sub-components: \textit{Causal Graph: Training Data} and \textit{Causal Graph: Test Data}, which visualize feature relationships in the training and test sets, allowing participants to click on a feature to highlight its connected edges; \textit{Feature Importance} highlights the AI model's key influencing features, and \textit{Data Distribution} displays the feature distribution within the training set.
\textbf{Component D} allows participants to select and rank features by using checkboxes and dragging them into their preferred order. Clicking \textbf{Confirm Feature Selection \& Ranking} navigates to Task 2. }
\label{fig: UIHomepage_Indicator}
\Description[Prototype System Dashboard]{Prototype system dashboard used in Task 1: feature selection and ranking. It shows the test dataset with feature values, model predictions, and explanations. Visualizations illustrate feature relationships, feature importance, and data distributions. The interface also lets participants select and rank features before proceeding to the next task.}
\end{figure*}

\begin{figure*}[htbp]
\centering
\includegraphics[width=0.85\linewidth]{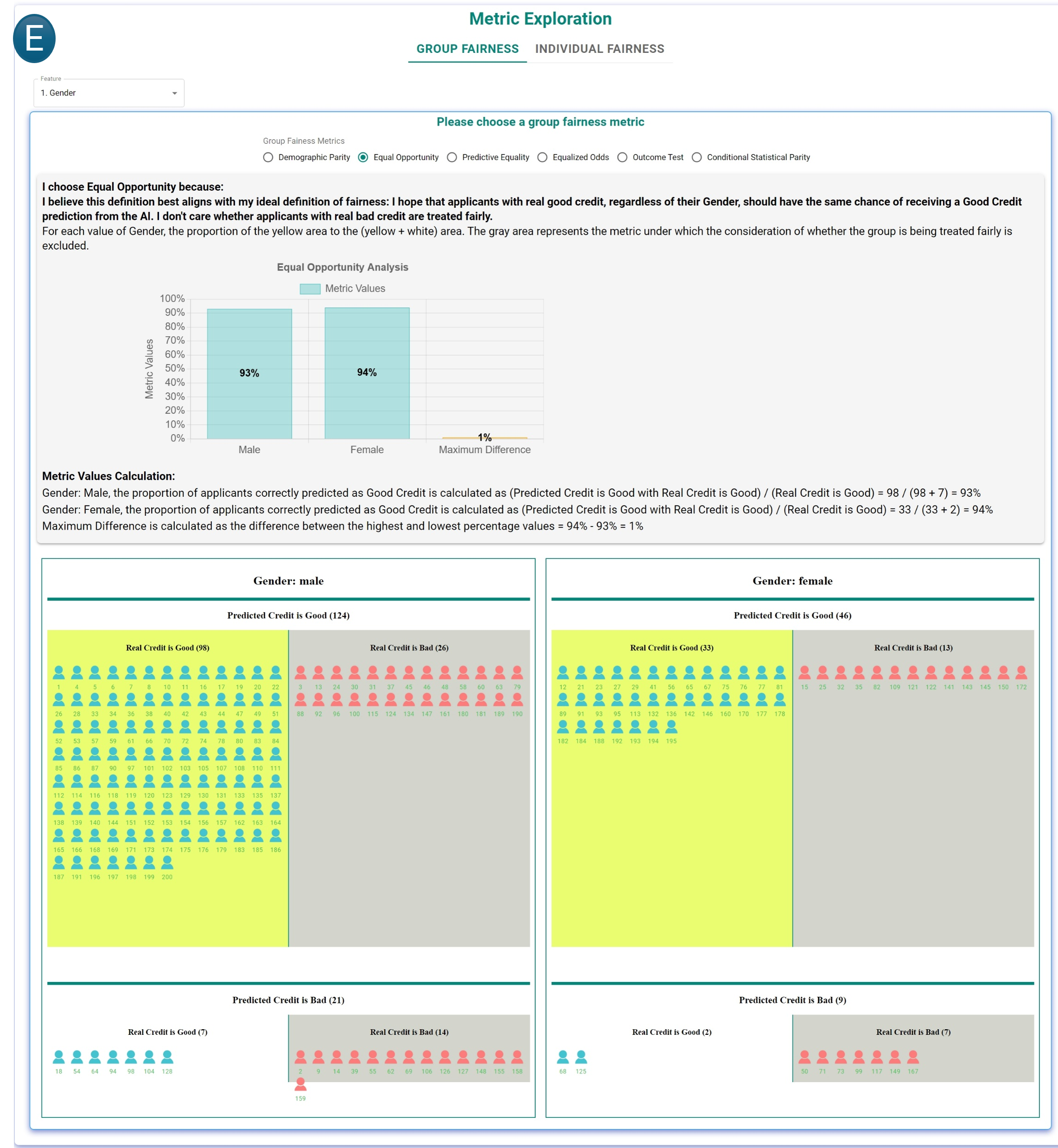}
\caption{\textbf{Prototype System: Feature-Metric Pairing Module (Task 2).} 
\textbf{Component E} shows a participant viewing (Group Fairness - Equal Opportunity) for Gender. The interface provides a lay definition, a bar chart with metric results, and metric calculation. Next, it shows a transparent individual-level visualization of the metric calculation, where \rev{the test dataset--all applicants--}are sub-grouped by gender and AI prediction (i.e., ``Predicted Credit is Good'' and  ``Predicted Credit is Bad''), and then further grouped by ground-truth label (``Real Credit is Good'' and ``Real Credit is Bad''). The gray region denotes applicants with real bad credit, and are excluded from this metric calculation. For applicants with real good credit, the metric is calculated as the proportion of Yellow (real good \& predicted good) over Yellow + White (all real good).}
\label{fig: Feature-Metric Pairing}
\Description[Prototype System Metric Selection]{Prototype system module for Task 2: metric selection. It lets participants pair a fairness metric with a feature, then shows the metric's explanations, visualizations, and detailed calculations.}
\end{figure*}
 
\subsection{Prototype System}
\label{sec:prototype system}
\subsubsection{Functionalities Supporting Study Tasks}
We designed the prototype system to support participants in the Main Phase tasks to assess AI outcome fairness. Building on prior work that highlights the value of contextual information for fairness-related decision making \cite{10.1145/3411764.3445308, 10.1145/3544548.3581227}, our prototype provided such information with metric explanations to help participants reflect on fairness. We outline its core functionalities, with technical details in Appendix~\ref{app:prototype system design}, and make the source code publicly available on GitHub.~\footnote{The
prototype source code is publicly available on GitHub: \url{https://github.com/LInnnLUooo/Lay-Stakeholder-AI-Fairness-Assessment-Tool}}

\textbf{Task 1: Feature Selection \& Ranking}
To support participants in completing Task 1, the system offered following functionalities:
\begin{itemize}
    \item Participants could use the prototype system to explore \rev{each case (i.e., each applicant) from the test dataset}, the model, and the features with their descriptions used for the model training (Figure~\ref{fig: UIHomepage_Indicator}). They were able to investigate \rev{each applicant's feature values along with their} ground-truth label and the AI's predicted outcome (Figure~\ref{fig: UIHomepage_Indicator} A). 
    \item Participants could access plain-language explanations of how the model works and what protected features mean in AI fairness. (Figure~\ref{fig: UIHomepage_Indicator} B). 
    \item Inspired by prior work showing that lay stakeholders benefit from transparency to support comprehension and fairness reasoning \cite{10.1145/3313831.3376447, doi:10.1080/10447318.2022.2067936, 10.1145/3514258}, participants could explore feature information (Figure~\ref{fig: UIHomepage_Indicator} C) through causal graphs showing feature relationships in the training and test data (generated with CausalNex \cite{10.5555/3327546.3327618}); model-based feature importance, and training data distributions. These elements acted as ``white-box'' explanations, exposing the model's inner workings and training context \cite{10.1145/3290605.3300789}. Importantly, in our settings, these visualizations were not intended to steer participants toward particular features but to situate them in context. We also reminded participants that such information may itself reflect biases in the dataset, model, or labels, underscoring its role as transparency rather than prescriptive guidance. Participants were free to use or disregard them, combining system-provided insights with their own experience.
    \item Participants were able to communicate their decisions by selecting the features using checkboxes and dragging them to rank them (Figure~\ref{fig: UIHomepage_Indicator} D).
\end{itemize}

\textbf{Task 2: Mapping Features to Metrics with Thresholds}
To support Task 2, our system helped participants build a clear understanding of fairness metrics, so they could express their fairness concerns in relation to specific features. We drew on the Fairness Explainer and Explorer (FEE) framework \cite{earnfairness}, originally designed to empower stakeholders without AI expertise in exploring outcome fairness metrics, and extended it in several ways.

Similar to FEE, participants could use both group fairness and individual fairness metrics: five group fairness metrics (Demographic Parity \cite{10.1145/2090236.2090255, NIPS2017_a486cd07}, Equal Opportunity \cite{10.5555/3157382.3157469}, Equalized Odds \cite{10.5555/3157382.3157469}, Predictive Equality \cite{10.1145/3097983.3098095}, Outcome Test \cite{articleoutcometest}, Conditional Statistical Parity \cite{10.1145/3097983.3098095}), and two individual fairness metrics (Counterfactual Fairness \cite{NIPS2017_a486cd07} and Consistency \cite{pmlr-v28-zemel13}). Definitions introduced in Section~\ref{sec: Related Work} with formula calculations provided in Appendix~\ref{app:Fairness Metrics and Explanations}. For each metric, participants could access multiple forms of explanation: a plain-language metric description, detailed metric calculations, and instance-level transparency based on ground-truth labels (``Rated Credit''), AI predicted outcomes (``Predicted Credit''), and feature values.

Unlike prior systems \cite{earnfairness, 10.1145/3411764.3445308} which were designed around AI experts' norms for assessing fairness, our prototype supports 1) participants to apply fairness metrics to any feature they considered relevant and 2) to help make metrics accessible and understandable. 
Beyond technical definitions, we also included descriptions of each metric's strengths and limitations, informed by feedback from lay stakeholders \cite{earnfairness, 10.1145/3411764.3445308}. These descriptions aimed to reduce the cognitive burden of metric choice, make selection more contextual, and reflect lay stakeholders' preference for understanding the practical implications of fairness metrics \cite{earnfairness}, thereby supporting participants in aligning metrics with their own values. Further, we introduced a “Custom Fairness” option that allowed participants to define their own metrics when predefined metrics did not capture their fairness concerns. Given the lack of prior work on lay stakeholders' custom metric design, we implemented this as an open-ended form: participants verbally expressed definitions while researchers documented and probed for clarification; participants could also handwrite their definitions if preferred.

Participants were able to make a pairing of features and metrics by selecting a feature from their ranked list and using a dropdown menu to select an associated metric. Figure~\ref{fig: Feature-Metric Pairing} shows how a participant could first select \textit{Group Fairness} and then choose the feature \textit{Gender} and \textit{Equal Opportunity} metric to view a textual explanation, bar-chart visualizations of the metric result, detailed metric calculations, and transparent groupings at the individual level. Participants were able to view all of these explanations together or focus on any single one they find most helpful. Through our prototype's presentation of metric explanations and results visualizations, participants also reflected on thresholds for selected feature–metric pairs, numeric boundaries beyond which they perceived the AI system as unfair. Participants articulated their fairness thresholds while interacting with the prototype, drawing on their own lived experiences and contextual reasoning.

\textbf{Task 3: Fairness Re-Ranking}
To support Task 3, our system allowed participants to revisit their fairness concerns by showing their previously selected feature–metric pairs, which they could then re-rank through a drag-and-drop interface if they chose to.

\subsubsection{Comparison with Existing Tools}
We detailed the design requirements identified from people without AI expertise \cite{doi:10.1080/10447318.2022.2067936}, and compared our prototype system with other common tools in fairness assessments also designed for lay stakeholders: FET \cite{10.1145/3411764.3445308}, FairHIL \cite{doi:10.1080/10447318.2022.2067936}, FEE \cite{earnfairness}, highlighting how each tool addressed the needs of participants without AI expertise (Table~\ref{tab:UI comparison}). Compared to existing tools, our prototype addresses multiple design requirements and further extends functionality through feature selection, feature-metric pairing, and priority-based ranking.


\begin{table*}[ht]
  \caption{Tool Design Comparison between FET, FairHIL, FEE, and Our Prototype System.}
  
  \label{tab:UI comparison}
  \centering
  \footnotesize

  \resizebox{\textwidth}{!}{%
  \begin{tabular}{p{2.5cm}lp{5.5cm}p{0.5cm}p{0.6cm}p{0.5cm}p{1cm}}
    \toprule
    \textbf{Area}&\textbf{Use}&\textbf{Requirement} &\textbf{FET} &\textbf{FairHIL} & \textbf{FEE} & \textbf{Our Tool}\\
    \midrule
    1. Attribute overviews& Informational& 1.1. Attributes, number of records and attribute value distributions& \checkmark & \checkmark  & \checkmark & \checkmark\\
    & & 1.2. Amount of missing data& & & &\\
    & & 1.3. Fairness metrics for model and individual protected attributes&  \checkmark &  \checkmark & \checkmark & \checkmark\\
    & & 1.4. Target distribution&  &  \checkmark &  \checkmark &  \checkmark\\
    \multirow[t]{7}{*}{\raisebox{-1.6ex}{\makecell[l]{2. Investigate relationships\\ between attributes}}} & Informational& 2.1. Distribution of protected attributes with other attributes&  & \checkmark &  &\\
    & & 2.2. Distribution of user-selected attribute values (e.g., job) and target values &  &  \checkmark  & \checkmark & \checkmark\\
    & & 2.3. Distribution of two user-selected attribute values (e.g., job) &  &  \checkmark  &  & \checkmark\\
    & Functional & 2.4. Support creation of new attributes (i.e., calculated from other attributes) &  & \checkmark  & & \\
    & & 2.5. Ability to create/include own fairness metric (if not already in the system) &  & \checkmark  &  & \checkmark\\
    & & 2.6. Allow creation of subgroups based on a combination of attributes and see their distribution on target & \checkmark & \checkmark  & \checkmark &\\
    & Adjust model& 2.7. Input custom thresholds to affect AI model& &  & \checkmark &\\
    3. Individual cases& Informational & 3.1. Specific application and attribute values& \checkmark & \checkmark  & \checkmark & \checkmark\\
    & & 3.2. Fairness metric for individual case &  &  & \checkmark & \checkmark \\
    & & 3.3. Level of similarity between cases & \checkmark & \checkmark  & \checkmark & \checkmark\\
    & & 3.4. Select specific cases to compare and show which attributes are similar & \checkmark & \checkmark & \checkmark & \\
    & & 3.5. Show decision boundaries &  & \checkmark & \checkmark  & \checkmark\\
    & Functional & 3.6. See “What If” results on target based on changes to attribute values &  &  & \checkmark  & \checkmark\\
    4. Model & Informational & 4.1. How model works &  & \checkmark  & \checkmark & \checkmark \\
    &  & 4.2. How it was created, rationale for decisions in modeling &  &  & \checkmark & \checkmark \\
    &  & 4.3.  Who created it &  &  & \checkmark & \checkmark\\
    &  & 4.4.  Explanations of when measures indicated unfairness or discrimination &  &   & \checkmark  & \checkmark\\ 
 \bottomrule
\end{tabular}
}
\end{table*}

\subsection{Data Collection and Analysis}
\label{sec: data collection and analysis}
We collected screen recordings of each participant's interactions with the prototype system and audio recordings of their concurrent think-aloud sessions. All audio was transcribed verbatim and lightly cleaned for readability. Screen recordings were used to corroborate transcript accuracy, resolve ambiguities in participants' utterances, and provide a trail of interaction steps that participants took through feature selection, feature–metric pairing, and ranking.

Because our research specifically focuses on the outcome fairness perspective, our data analysis draws on data from 26 out of 30 participants who assessed fairness through this perspective. To maintain conceptual clarity and a coherent outcome fairness assessment, we excluded all collected data that did not reflect an outcome fairness assessment, i.e., complete data from the remaining 4 participants who held the \textit{affordability fairness} perspective \cite{10.1145/3514258}, and the partial data reflecting the \textit{fairness through unawareness} notion \cite{NIPS2017_a486cd07, hort2023bias} from 3 of the 26 participants who otherwise held an outcome fairness perspective. These excluded perspectives focused on feature usage during `model training': caring about which features should be included (affordability fairness) or excluded (fairness through unawareness). When participants assessed fairness through these perspectives, they were not evaluating whether the model's prediction outcomes were biased with respect to those features, nor were they engaging with answering: whom to protect, how to protect, and to what extent. Although we noted that participants with different perspectives sometimes raised the same features (e.g., Gender), indicating concern about the same type of bias, they were reasoning and evaluating fairness from fundamentally different stages of the model pipeline. Table~\ref{tab: Participant include and exclude} in Appendix~\ref{app:Participant Information} provides an overview of participants whose data were included or excluded.

We conducted a primarily qualitative analysis of the transcribed data. First, two researchers reviewed the transcripts to extract the features participants selected, their original and final rankings, and the mappings between features and fairness metrics, including corresponding fairness threshold settings. For participants who designed custom metrics, we supplemented this data with detailed session notes from both researchers and participants to capture the structure, properties, and intended rules for applying those metrics. We also performed a thematic analysis \cite{doi:10.1191/1478088706qp063oa} to identify participants' reasoning behind their feature selection and metric mappings. 

We coded the transcripts and refined the codebook through five iterative cycles involving multiple researchers. After each cycle, the coding was jointly reviewed in team meetings to resolve differences, clarify definitions, and reach consensus, ensuring that the final themes were grounded in the data. The final codebook is provided in Appendix~\ref{app:feature_selection_reasons_codebook}. 

Although our analysis was qualitative in nature, certain decision outputs (e.g., selected features, rankings, metrics, threshold values) were discrete and countable. We therefore report simple descriptive statistics to provide a high-level overview of observed patterns, such as frequencies, minimum and maximum, mean, mode, and standard deviations of selections, as well as counts of participants making specific choices.

\section{Results}
\label{sec: Results}
We first separately report the fairness perspectives participants expressed other than outcome fairness for completeness. We then answer our research questions (RQ1 and RQ2) to uncover how participants assessed AI outcome fairness. Finally, we describe participants' user experience during the assessment, including the challenges and benefits they encountered and the prototype system's SUS score.

\subsection{Other Fairness Perspectives}
\label{sec: different perspectives}
Only a few participants (4 participants; P5, P8, P13, P14) held the perspective of affordability fairness \cite{10.1145/3514258}, aiming for fairness by basing AI decisions on an individual's actual ability to afford a loan. As P13 emphasized, ``\textit{I will care more about this applicant's creditness. Um. This includes [...] like some features to see whether they can pay back or should get the loan or not}''. Similarly, P5 mentioned, ``\textit{Because the main objective of the bank is not just to do the things on the basis of morality only}''. Therefore, this perspective led them to select highly context-specific features that are considered reasonable and precise for evaluating an individual's credit level, like `Credit Amount', `Existing Credits', and `Checking Account', while deliberately disregarding demographic characteristics such as `Gender', which seem unrelated to credit evaluation but can introduce bias into the prediction. Among these four participants, two participants highlighted the importance of procedural fairness, emphasizing that all individuals should undergo the same evaluation process when assessing affordability. As P5 explained, ``\textit{If all the persons go from the same litmus paper, [using] same features, same characteristics, so it is, I think, fairer than fair. It is fair and transparent}''. Taken together, these four participants approached fairness by reasoning about how the AI model should be trained, that is, which features it ought to use to accurately capture people's affordability, rather than, in terms of whether the AI system's outcomes were fair for the groups or features they sought to protect.

Additionally, three participants (P9, P22, P25) among the 26 participants who held an outcome fairness perspective also mentioned fairness through unawareness fairness notion \cite{NIPS2017_a486cd07, hort2023bias}, explicitly expressing their preference for excluding features such as `Telephone', `Gender', `Foreign Worker', and `Residence Length' when training the AI model, as they viewed these features as irrelevant to the AI's decisions. 

\subsection{How do stakeholders identify features that are crucial for assessing AI outcome fairness, and how do they prioritize them? (RQ1)}
\label{sec: Ranked Feature Selection List after Trade-Offs}
Our study uncovered that selecting features for fairness assessment of AI systems involved complex decision-making for participants. We also explored the underlying reasons for their choices through a thematic analysis and the themes that were extracted are indicated in \textbf{bold text}. The remainder of our analysis focuses on the 26 participants who assessed outcome fairness. Accordingly, all subsequent analyses, including feature selection and ranking, metric selection, and threshold setting, are drawn from these 26 participants. The full results of these 26 participants are shown in Appendix~\ref{app:results table}.

\begin{figure*}[ht]
    \begin{subfigure}[t]{0.45\linewidth}
        \includegraphics[width=\linewidth]{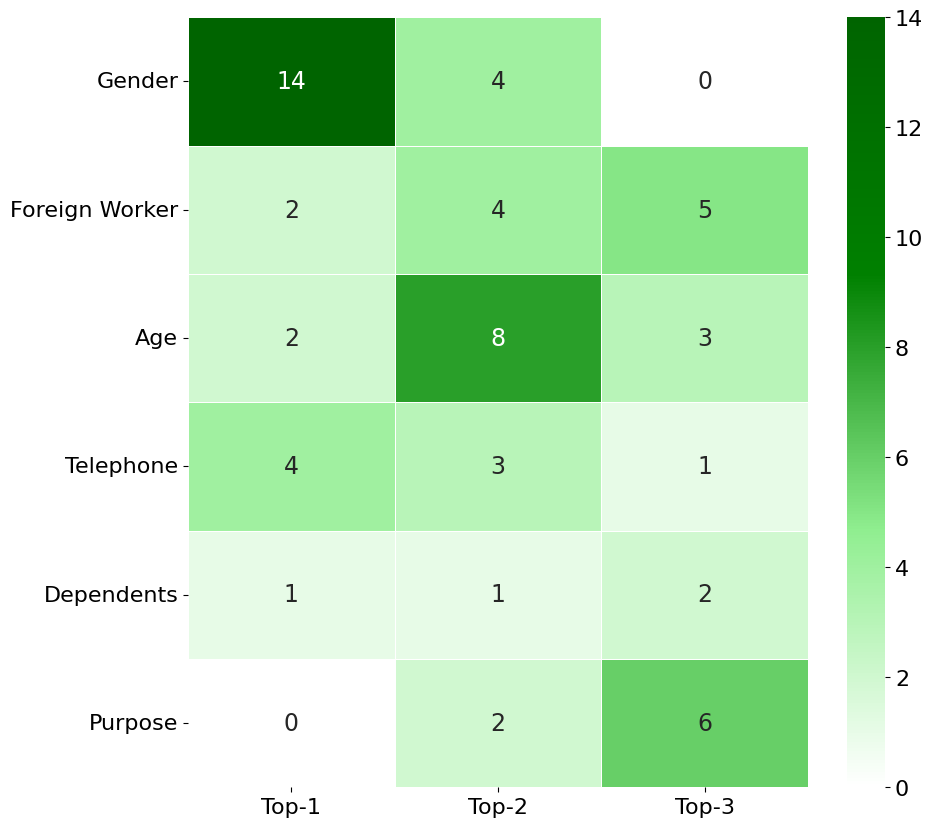}
        \caption{Top 6 most selected features and their selection frequencies in the Top 3 ranking positions.}
        \Description{Heatmap showing the six most frequently selected features and how often they appeared in participants' top three rankings.}
    \end{subfigure}
    \hfill
    \begin{subfigure}[t]{0.45\linewidth}
        \includegraphics[width=\linewidth]{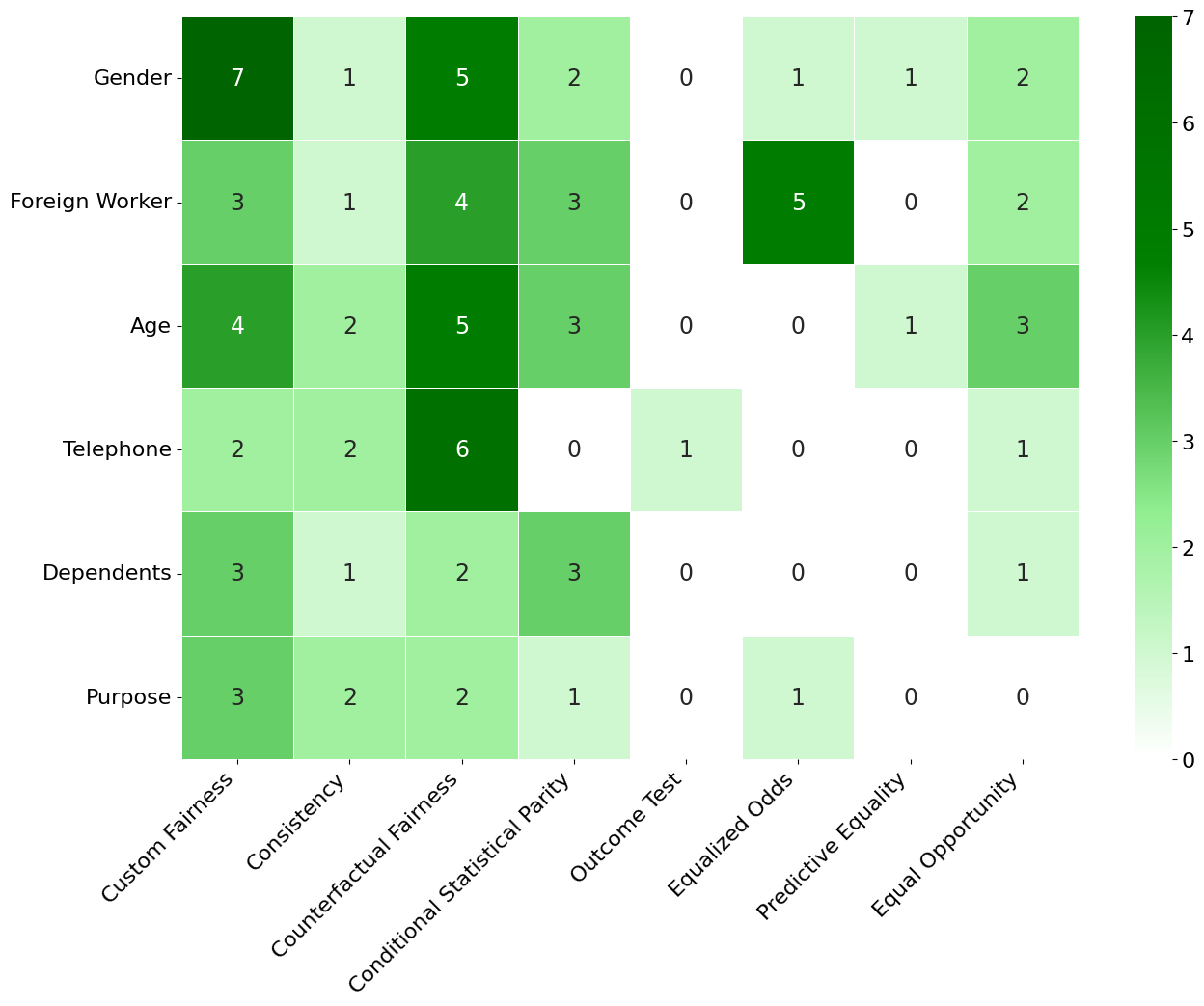}
        \caption{Top 6 most selected features and the distribution of their paired fairness metrics.}
        \Description{Heatmap showing the six most frequently selected features and the distribution of fairness metrics paired with each feature.}
    \end{subfigure}
    \caption{Heatmaps: Distribution of participants' top-selected features and their associated fairness metrics.}
    \label{fig: heatmaps}
    \Description{The figure on the left is a heatmap showing the six most frequently selected features and how often they appeared in participants' top three rankings. The figure on the right is a heatmap showing the six most frequently selected features and the distribution of fairness metrics paired with each feature.}
\end{figure*}

\subsubsection{Broad Fairness Considerations}
We observed that participants, when assessing outcome fairness, selected features that associated with \textit{whom they want to protect}, i.e., features they ``\textit{don't want this AI to discriminate against}'' (P30), such as `Gender', which could disproportionately harm or benefit certain individuals or groups. At the same time, they avoided selecting those highly context-specific features, such as `Credit Amount', which they considered reasonable and precise for producing prediction outcomes.

We found that participants demonstrated broader fairness considerations in outcome fairness than AI experts typically do. Previous work \cite{FairnessDatasetSurvey} has shown that for the German Credit Dataset, AI experts often select only three features: two protected features (`Age' and `Gender') and one sensitive feature (`Foreign Worker'). In other credit scenarios \cite{de2025towards}, AI experts' attention similarly tends to remain on a narrow set of protected or sensitive features (typically centered on gender and race), with relatively little attention paid to other potential sources of bias. In contrast, 19 of the 20 available features were selected at least once by participants, except for `Credit Amount'. In average, our participants selected six features for this credit rating scenario's fairness assessment (Mean = 6.15, Min = 1, Max = 11, Mode = 6, SD = 2.30), with the most frequently selected features overall being `Gender' (21 participants), `Foreign Worker' (21 participants), `Age' (20 participants), `Telephone' (15 participants), `Dependents' (12 participants), and `Purpose' (10 participants). These broad choices underscore the wide variation in how stakeholders interpret fairness, extending far beyond the narrow sets of protected features typically considered in technical fairness \cite{de2025towards, FairnessDatasetSurvey}. 

\subsubsection{Protected Features' Role: Central Yet Contested}
Our analysis showed that protected and sensitive features played a major role in participants' fairness assessments, but participants did not agree on their priority for fairness assessment. `Gender', `Foreign Worker', and `Age' were the top three most frequently selected features overall, and these three features also dominated the top three ranking positions (i.e., Top-1, Top-2, and Top-3), as shown in Figure~\ref{fig: heatmaps}(a), with `Gender' chosen by 18 participants mainly in the Top-1 position, `Age' by 13 participants mainly in the Top-2 position, and `Foreign Worker' by 11 participants mainly in the Top-2 and Top-3 positions. However, not all participants prioritized these protected and sensitive features in their fairness assessment. Only 18 out of 26 participants placed any of these features in their Top-1 position. Additionally, not everyone included any of these features in their selection at all, for example, `Gender' and `Foreign Worker' were omitted by 5 participants, and `Age' was excluded by 4 participants. 

In their choice and ranking of protected and sensitive features, many participants stated that they believed such features were \textbf{Not relevant to context-specific decision-making}, i.e., not related to loan repayment (16 participants). As P27 said, ``\textit{we shouldn't really be discriminating based on age, it should be based on whether or not they can pay back the loan [...]}''. Participants also expressed concerns about \textbf{Discrimination against inherent human traits and social labels} (12 participants). As P23 explained, ``\textit{I think the first [selected] ones [Age and Gender] are all inherent human traits. Um. Although the second one [foreign worker] doesn't fit perfectly, they all feel like labels that people naturally carry. If people judge others based on these, it feels a bit like discrimination, like looking at others through tinted glasses. So I rank them higher}''.

\subsubsection{Unprotected Features' Role: Complementary Yet Indispensable}
Our analysis also showed that non-protected features play a complementary yet indispensable role in participants' fairness assessments, a dimension often overlooked from a technical fairness perspective by AI experts \cite{de2025towards, FairnessDatasetSurvey}. Nearly all participants (25 out of 26) selected a combination of both protected and non-protected features to create a more holistic fairness assessment. Even though they are ranked generally lower than protected and sensitive features, nonetheless, they appeared frequently in the Top-3  position. As Figure~\ref{fig: heatmaps}(a) shows, `Telephone' was ranked highly by 8 participants, sometimes even appearing in Top-1 position, `Purpose' by 8 participants often appearing in the Top-3 position, and `Dependents' by 4 participants. Many non-protected features, such as, `Job', `Credit History', were also selected but ranked much lower, associated with context-specific reasons such as \textbf{Financial ability considerations} (19 participants). 

As reasons for their importance and thus their high ranking, participants reported that they were concerned that these non-protected features \textbf{Don't capture context-relevant information}, potentially leading to inaccurate assessments and introducing unfairness (9 Participants). As P3 said, ``\textit{the whole telephone thing, I think that's just nonsense. Whether someone leaves a phone number or not has just become another data point?}'' [P3]. Additionally, participants highlighted AI models should \textbf{Avoid prejudging people's choices} introduced by certain non-protected features when making decisions (8 Participants). As P19 mentioned, ``\textit{'Purpose' I think [...] we cannot prejudge, is that purpose good or bad or is that superior than any other purpose. So, everyone have their own choices}''.   

\subsubsection{Shifting Priorities: The Dynamic Nature of Fairness Decisions}
As they engaged with the fairness assessments, we observed shifts in feature priorities for 10 out of 26 participants, indicating a dynamic nature to their decision-making. Nine participants re-ranked features within their existing lists, while one participant (P16) made a substitution, replacing the unprotected feature `Credit History' with the protected feature `Age'. Further, we observed that participants who re-ranked features tended to make small adjustments, primarily to the top three important features. Among the nine participants, eight moved only one feature, with seven of these changes occurring within the top three features. When participants made these small adjustments, they often prioritized non-protected features (5 participants). For example, P22 re-prioritized `Dependents', and P28 prioritized `Purpose', moving these into the top three. In contrast, only three participants moved protected or sensitive features higher, such as P2 elevating `Foreign worker' above `Telephone' and P19 prioritizing `Age' over `Purpose'. The feature priority adjustments were driven by concerns of \textbf{Less contextual relevance} (7 participants) of certain features to the AI prediction outcome, aiming to reduce the potential biases these features might introduce into the results. As P16 mentioned, ``\textit{The only thing I would change is that is the credit history and age. I think they should be replaced because age just, age really does not matter [for the context], but credit history matters. A little bit more than age}''. 


\vspace{1em}
\noindent
\colorbox{bgcolor}{
  \parbox{\dimexpr\linewidth-2\fboxsep\relax}{
    \textbf{Findings summary}: Participants demonstrated broad and complex fairness concerns when selecting features to assess whether the model's outcomes discriminate based on those features. They tended to go beyond legally protected features and considered the potential biases introduced by non-protected features. This meant more features were attended to, and thus prioritized. Prioritizations were not static; participants changed how they ranked features as they engaged more deeply in fairness assessments.
  }
}

\subsection{How do stakeholders make metric choices to best reflect their perceived fairness for different features, and how do they set fairness thresholds? (RQ2)}
\label{sec: Metric Selection Preferences and Reasons}

From the 26 participants who had concentrated on outcome fairness, we obtained a total of 128 feature-metric pairs along with their fairness thresholds. (For full results, see Appendix~\ref{app:results table}.) We analyzed these results to uncover the complexities and nuances in stakeholders' approaches to choosing metrics and pairing them with features.

\subsubsection{Misalignment in Metrics: Stakeholders' Diverse Needs vs. Technical Practice}
We investigated the overall metric preferences based on the eight predefined metrics and the `Custom Fairness' metric option. The three most chosen metrics were Counterfactual Fairness, selected 31 times across all participants, Custom Fairness (28 times) and Equal Opportunity (20 times). Less frequently selected metrics were Conditional Statistical Parity (18 times), Consistency (18 times), and Equalized Odds (10 times). This highlights that participants' fairness needs were diverse, spanning both individual and group fairness categories. Participants also appeared to frequently select more nuanced and complex metrics such as counterfactual fairness and custom fairness, citing the necessity of assessing individual circumstances and the requirement for flexible defining fairness. 

In contrast, AI experts often choose group fairness metrics that are easier to implement and widely used in legal or compliance settings \cite{conflict/10.1145/3351095.3372864, 10.1145/3287560.3287598}, such as Demographic Parity. Prior work commonly recognizes Demographic Parity as one of the most frequently used fairness metrics, both within credit-related contexts \cite{de2025towards} and across other AI applications scenarios \cite{10.1145/3457607, 10.1145/3194770.3194776, hort2023bias}. However, this metric was not selected by any of our participants. Similarly, individual fairness metrics such as Counterfactual Fairness and Consistency, as well as more nuanced and more complex-to-implement group fairness metrics like Conditional Statistical Parity, were popular among participants but appear to be used less often by AI experts \cite{de2025towards, hort2023bias, 10.1145/3457607, 10.1145/3194770.3194776}. This difference in metric preferences highlights a potential misalignment between current technical fairness practices and human-centered fairness expectations.

\subsubsection{Context Matters: Stakeholders Switch Metrics Across Features}
Next, we analyzed each participant's metric selections. We found that 20 participants often switched metrics for different features. For example, 10 participants used 3 different metrics, 8 used 2 metrics, and 2 used 4 metrics. An illustrative example is P7, who ranked and evaluated four distinct features, each aligned with a unique metric: (Consistency | `Credit History'); (Custom Fairness| `Purpose'); (Equal Opportunity | `Foreign Worker'); (Conditional Statistical Parity | `Gender'). We observed that participants seemed to first consider their preferred fairness category, i.e., individual or/and group fairness, and then decided on specific metrics within these categories. For example, P16 explained, ``\textit{so to me the group fairness doesn't take that much priority because the individual fairness is what matters}'' which then led to pairing different metrics within the individual fairness category with the selected features. Only six participants followed a ``one-size-fits-all'' metric selection approach, similar to the practice often adopted by AI experts across both credit scenarios and other AI applications scenarios \cite{de2025towards, hort2023bias}, where they chose the same metric (or metric combinations under Custom Fairness) across all of the features they had selected. 



\subsubsection{Metrics for Features, and the Reasons Behind Them}
We found that each of the top ranked features (`Gender', `Foreign Worker', `Age', `Telephone', `Dependents', and `Purpose') were paired with multiple metrics: at least five different metrics in the case of `Telephone', `Dependents', and `Purpose', to seven metrics for `Gender'. Since participants chose different metrics for different features, we investigated any patterns within feature-metric pairings, concentrating on the six most selected features and their most frequently selected metrics (Figure~\ref{fig: heatmaps}(b)).

We found that Counterfactual Fairness was most frequently selected for `Age' (5 participants) and the unprotected feature `Telephone' (6 participants). Recall that the Counterfactual Fairness metric measures whether an individual receives the same AI outcome when certain feature values are hypothetically altered while all other features remain unchanged. Underlying this choice was often that they wanted to \textbf{Remove the influence of irrelevant features} (8 participants) when assessing AI fairness. For example, P22 highlighted for `Age', ``\textit{They [referring to the system] changing the age, and that shouldn't change things. I think in this case it's a question of treating people the same}''.

Equalized Odds was the most selected metric for `Foreign Worker' (5 participants), which measures both Equal Opportunity (ensuring protected and unprotected groups have the same probability of receiving favorable outcomes when they truly qualify) and Predictive Equality (ensuring both groups have the same probability of incorrectly receiving favorable outcomes when they do not qualify). Participants often chose this metric because they focused on ensuring \textbf{Disparity controlled across all parties} (5 participants), aiming for \textit{``more comprehensive''} (P28) fairness across groups rather than limiting their focus to fairness within qualified (measure by Equal Opportunity) or unqualified (measured by Predictive Equality) groups. As P22 explained, ``\textit{[...] applicants [...] regardless good credit or bad credit, should have the same chance of receiving good credit. Just have the same chance. A fair chance of getting it at the same rates as people in the native population}''.  

Conditional Statistical Parity emerged as one of the most selected metrics for `Dependents', chosen by 3 participants because they wanted to customize groups and perform \textbf{Feature-specific comparisons}. As P25 explained, ``\textit{if I was the bank owner okay and I saw that this person has a high dependence okay I should say that I need to check another thing, the higher amount of salary or savings. We need to consider another factor}''. Among them, two participants also mentioned a demand for a more \textbf{Objective approach to grouping people} for group-based fairness assessments, especially when they believe that the feature being evaluated for fairness has a significant influence on other features. As P28 said, ``\textit{Dependents may affect other factors. There may be a relationship between them I think it still depends on realistic features that it might influenced}''. Participants believed that comparing groups based on specific features and their values is more objective than using groups solely defined by ground truth labels, as it more accurately reflects individuals' actual abilities and characteristics.

Finally, we found that when predefined metrics did not meet participants' perceptions, they were trying to define their own fairness metrics. Custom Fairness emerged as one of the most selected metrics for `Gender' (7 participants), `Dependents' (3 participants), and `Purpose' (3 participants). Participants for the most part tried to combine different existing metrics to achieve a more comprehensive fairness assessment, for example by combining ``Consistency and Counterfactual Fairness'' for `Gender' (2 participants) or ``Equal Opportunity and Counterfactual Fairness'' for `Gender' (2 participants), `Dependents' (2 participants) and `Purpose' (2 participants). 

In constructing these custom metrics, such as ``Consistency \& Counterfactual Fairness'', participants wanted to meet \textit{both} individual fairness metrics \textit{simultaneously}. This, in their mind, would balance \textbf{Comparison with similar individuals} and \textbf{Remove the influence of irrelevant features}. Participants also mentioned the similarity between these two metrics. As P16 explained:
    ``\textit{I think it's interesting because I feel like the two things are connected so I felt like if counterfactual fairness talks about having the same result even when those specific features changed, and then consistency talks about between the between any two candidates they should have the same result because they are similar. [...] I felt like if we do one of these things another metric will also come true}''. 

A different way to fashion a Custom Fairness metric can be seen in ``Equal Opportunity \& Counterfactual Fairness'', where different aspects of metrics are combined. Participants who chose this combination aimed to \textbf{Remove the influence of irrelevant features} on AI outcomes through the individual-level Counterfactual Fairness while also trying to inject group-level comparisons with Equal Opportunity by \textbf{Focusing only on qualified individuals}, thus ensuring fair treatment for individuals with a ``Good Credit'' ground-truth label. As P20 highlighted, ``\textit{So for `Dependents', I would like a situation of counterfactual fairness, because I think for each individual, like when their dependents [number] change, the AI model should always give them the same prediction, and then as well as that [equal opportunity], for the people that can actually have good credit, that can actually afford it [can be treated fairly]}''. In contrast to the previous instance where two metrics within the same fairness category are simply assumed to be optimized at the same time, this cross-category combination functions more like a new fairness metric, integrating desirable elements from each metric and potentially posing greater challenges during mitigation.

\subsubsection{Diverse but Strict Thresholds}
\label{sec: thresholds}
An important part of fairness decision-making is how much unfairness is acceptable, 
which we measured using fairness thresholds. Recall that we defined fairness thresholds as the maximum allowable deviation from the optimal fairness outcome of the current metric. Participants set thresholds ranging from 0\% to 100\%, with higher thresholds indicating greater tolerance for unfairness. Looking at the top Feature-Metric pairings we just described, we found that participants' fairness demands were notably stringent for existing metrics, while still allowing for \textbf{Some room for unfairness} (13 participants). For example, thresholds were set at on average 6.00\% for (Counterfactual Fairness|'Age') and 4.75\% for (Counterfactual Fairness|'Telephone') while this is even lower for some feature-metric pairings, such as 3.20\% for (Equalized Odds|'Foreign Worker'), and 3.50\% for (Conditional Statistical Parity|'Dependents'). For Custom Fairness, there are no easy ways to compare between threshold levels but we found that they fluctuated a lot, for example from 0.5\% for (``Counterfactual Fairness \& Equal Opportunity''|'Gender') to 40\% for (``Outcome Test \& Counterfactual Fairness''|'Purpose') but these might be due to individual differences rather than a consistent pattern. However, the thresholds specified by our participants usually exceeded thresholds often used by AI experts, such as the legal benchmark of 20\% deviation \cite{10.1145/2783258.2783311} or a common AI development standard of around 10\% deviation \cite{Gupta2021TransitioningFR}, regardless of whether the features are protected or non-protected.

\vspace{1em}
\noindent
\colorbox{bgcolor}{
  \parbox{\dimexpr\linewidth-2\fboxsep\relax}{
    \textbf{Findings summary}: In feature-metric pairings, participants tended to choose different metrics for features rather than relying on a ``one-size-fits-all'' metric selection approach. They also showed a strong interest in creating more complex custom metrics, either to ensure that two or more metrics would be met at the same time or by trying to integrate aspects of different metrics that they liked. Participants seemed to adjust their fairness thresholds based on specific feature-metric pairs, demonstrating high expectations for fairness and low tolerance for unfairness.
  }
}

\subsection{User Experience: Challenges and Benefits}
\label{sec: experience}
Our participants reported two main challenges in the process of fairness assessment. First, 9 participants noted that \textbf{understanding different fairness metrics and terms required considerable time and effort}. As one noted, ``\textit{It took me quite a while to understand the different types of fairness… it also felt that both individual and group fairness should be taken into account… It was also hard to remember the different types and compare at a glance}'' (P27). This challenge was not unexpected: participants had no AI expertise, yet assessing AI fairness required them to understand each metric and decide which one best fit their fairness assessment needs. Second, only 3 participants pointed out that \textbf{information overload} in the prototype made it ``sometimes difficult to concentrate on where to focus'' (P19). As P26 noted, ``\textit{We have many parameters and they are complexly interrelated. To make a decision, I need to understand them fully}''. 
 
These challenges highlight a tool design tension: providing enough information for stakeholders to understand and assess AI fairness without causing information or cognitive overload. To tackle these challenges, we observed that for metric exploration, participants typically moved through the interface from top to bottom: reading the textual definition, checking the visualization and calculation, and then understanding instance-level explanations. They deepened their understanding by selecting the explanation formats that suited them and lingering on the explanations to think them through. When deciding on a metric, participants often switched across UI pages to compare the pros and cons of different metrics, align the options with their needs, and consider whether to create custom fairness metrics. We will outline potential tool improvements in Section~\ref{sec: functionality and improving usability}.

While participants noted challenges, they all successfully understood and independently completed all three tasks without researcher intervention \rev{in their fairness decision-making}. Because participants thought aloud and the researcher asked follow-up questions when needed, their explanations made clear how they interpreted the scenario, understood different features and fairness metrics, compared different fairness metrics, and articulated why particular metrics aligned with their concerns for specific features. These real-time observations enabled us to verify participants' correct understanding, while our thematic analysis captures how they reasoned with this understanding to make value-driven fairness assessment choices.

Participants also provided positive feedback on our prototype system. First, eighteen participants appreciated the overall ``\textbf{flow of information}'' (P9), such as dashboard integration, and noted that the prototype was ``\textbf{easy to use}'' (P26) and ``\textbf{easy to navigate}'' (P29). Second, five participants valued the \textbf{intuitive and simplified presentation of complex concepts}, noting that ``\textit{the interaction between features exploration is super clear and direct to understand}'' (P24) and appreciating the ``graphics alongside fairness definitions'' (P4).

We also collected participants' System Usability Scale (SUS) \cite{SUSEvaluation} scores to quantitatively assess the usability of our current system (Appendix~\ref{app:Post Questionnaires}), thereby informing targeted directions for future design improvements. The overall average SUS score was 61.83, with nearly 75\% of participants rating the system between 60 and 100, suggesting moderate usability relative to the desirable benchmark of \textit{68} \cite{SUSEvaluation}. A closer look at item-level responses provides a more nuanced picture. The SUS includes both ``positive'' (odd) items and ``negative'' (even) items: for positive items (0–5 scale, higher is better), participants reported strong agreement that the system is easy to use (23/26), that information is well integrated (26/26), that they felt confident when using it (23/26), and that it is something they would like to use frequently for AI fairness assessment (24/26). Notably, all these items received scores of at least 3, reflecting solid usability signals. 

For negative items (0–5 scale, lower is better), very few participants perceived the system as complex (4/26), inconsistent (1/26), or cumbersome (2/26), and these scores fell between 1 and 3, indicating that negative perceptions were limited. While lower scores on questions about needing support contributed to reducing the overall SUS score, this was expected since the system was designed to be used \rev{with researcher assistance for technical system usage, available when needed} during the study. Taken together, these results suggest that, despite the inherent complexity of fairness assessment tasks, participants generally found the \textbf{prototype system straightforward, confidence-enhancing, and well-integrated}, offering a strong foundation for further refinement.

\section{Discussion}
\label{sec: Discussion}
We acknowledge our study's limitations and outline future directions. We identify patterns in stakeholders' fairness assessment behavior and propose tool design implications to support stakeholders in assessing AI outcome fairness. We also provide possible approaches to manage the complexities of stakeholders' fairness decisions.

\subsection{Limitations and Future Work}
\label{sec: limitations}
Our qualitative study with 26 stakeholders without AI expertise uncovered rich and detailed insights into their behaviors and underlying reasoning when assessing AI outcome fairness. As an initial step in this direction, we acknowledge several limitations that also point to opportunities for future work. 

First, while focusing on a single stakeholder type allowed for more targeted analysis, the complexity of behavior is likely to increase when involving a broader range of stakeholder roles (e.g., policymakers, domain experts), backgrounds, and characteristics. Inspired by prior research on fairness perceptions \cite{doi:10.1080/10447318.2022.2067936}, participants' fairness assessment behaviors may also be shaped by their cultural and societal norms. Future research could expand our methodology with larger and more diverse participant samples, enabling an examination of how fairness assessment varies across roles and cultural contexts. Moreover, we observed the emergence of differences between lay stakeholders' (decision subjects) fairness assessment decision-making and the prevailing practices of AI experts as reflected in prior work \cite{de2025towards, hort2023bias, 10.1145/3194770.3194776, rabonato_systematic_2025}. Future work could involve both AI experts and lay stakeholders to collaboratively explore ways to reconcile differences and build consensus toward more inclusive AI fairness.

Second, our study was intentionally situated in the credit rating context, and although the German Credit dataset is real-world, it is relatively dated compared with more modern datasets that contain richer features, missing values, or multi-modal data. This might constrain the breadth of fairness assessment behaviors we were able to observe. Nonetheless, the underlying patterns of complexity and value-driven reasoning we uncovered in this scenario may still advance our understanding of stakeholders' needs when assessing fairness and offer useful insights for HCI-driven AI fairness design. Future research could extend this fairness assessment to diverse domains (e.g., healthcare, hiring, criminal justice) to surface context-specific nuances, as well as examine how decision-making patterns vary across scenarios with different levels of risk \cite{2024crowdsource/10.1145/3640543.3645209}.

Third, although stakeholders expressed clear preferences for customized fairness metrics, the current prototype system does not support direct custom fairness input in an operational formulation, thereby providing limited insight into operationalization and mitigation in practice. Future research should investigate tools and processes that not only help stakeholders articulate custom fairness definitions that reflect their expectations, but also translate these lay concepts into operationalizable metric formulations, thereby supporting the validation and integration of custom metrics into fairness assessment and mitigation workflows. 

Additionally, although we found other fairness perspectives held by our participants, our prototype was designed specifically for outcome fairness and did not support fairness assessments grounded in other perspectives, such as affordability fairness or fairness through unawareness, which required specifying which features to include or exclude during model development and evaluating fairness on that basis. Future work should investigate how stakeholders could be supported in articulating different fairness perspectives and examine what factors might shape these perspective differences, such as demographic factors.

\subsection{Patterns in Stakeholders' AI Fairness Assessment Decision-making}
\label{sec: patterns}
We supported participants in assessing AI outcome fairness through feature selection (access to both protected and unprotected features), ranking (to express fairness priorities), metric selection (with group, individual, and custom fairness options), and threshold setting (at the feature–metric level). As fairness is highly contextual \cite{10.1145/3711079} and changes in domain or even dataset can lead to different selection outcomes, we therefore do not focus on participants' specific choices. Instead, we carefully identify six recurring decision-making patterns that capture overall preferences in how participants assessed fairness across features, metrics, thresholds, and priorities. We next discuss these six patterns alongside prior research and reflect on how they may offer some tentative insights into how lay stakeholders assess AI outcome fairness.

\subsubsection{Broad feature selection with expanded fairness considerations} Although participants prioritized protected features such as `Gender', which resonates with prior user studies \cite{Saxena/10.1145/3306618.3314248, hort2023bias} and reflects alignment with established legal and ethical frameworks in the EU \cite{EUCharter2012}, the US \cite{USLaw}, and the UK \cite{EqualityAct2010}, they also elevated non-protected yet contextually less relevant (e.g., to financial ability in our credit rating scenario) features, such as `Telephone', or `Purpose', to high fairness relevance. These observations lend partial support to previous hypotheses that non-protected features might carry fairness weight in high-stakes scenarios \cite{earnfairness, NonprotectedFeatures, conflict/10.1145/3351095.3372864}. Taken together, our findings imply that when stakeholders identify fairness-related features, they are likely to consider contextual relevance and the potential bias a feature may introduce, rather than relying on legally protected characteristics. By contrast, AI experts' current practice in credit contexts \cite{de2025towards} and other AI domains \cite{10.1145/3641276} often relies on a limited set of legally protected features. The broader feature considerations shown by stakeholders suggest that this area may warrant more attention.

\subsubsection{Preference for fine-grained fairness assessment} From participants' overall metric choices, we observed a preference toward fine-grained, comprehensive fairness measurement, in contrast to coarse group-based parity. This preference was reflected, in part, through participants' frequent selection of Counterfactual Fairness, which they described as a way to treat people fairly at the individual level and to remove the influence of irrelevant features when AI makes decisions. In contrast, coarse group-based measures were relatively unpopular. For example, no participant selected Demographic Parity. When participants pursued group fairness for features, they leaned toward more fine-grained forms of group assessment. For example, more people preferred metrics like Conditional Statistical Parity for customizing comparison groups and making feature-specific, refined group evaluation, compared to Demographic Parity or Equalized Odds. This pattern resonates with similar metric preferences in prior studies involving lay stakeholders in credit rating, loan decision-making, and child maltreatment risk prediction, which also showed that stakeholders examined fairness at a more fine-grained level rather than relying solely on coarse group statistics \cite{Saxena/10.1145/3306618.3314248, doi:10.1080/10447318.2022.2067936, earnfairness, 10.1145/3411764.3445308}. Recent work in personnel selection also reports a similar preference for 'concrete' fairness: stakeholders often felt the need to supplement group-based fairness metrics with more individual-level examinations, noting that coarse group statistics could overlook important nuances \cite{10.1145/3757647}. In contrast, current practice among AI experts tends to simplify fairness assessment to group-statistical comparisons, such as Demographic Parity, because they offer greater operational practicality, align with legal requirements, and facilitate comparability, especially during bias mitigation \cite{de2025towards, hort2023bias, 10.1145/3194770.3194776, 10.1145/3457607, conflict/10.1145/3351095.3372864}.

\subsubsection{Favoring custom metrics} Given that people may hold different expectations of fairness shaped by their lived experiences and backgrounds \cite{landers2023auditing,10.1145/3334480.3375158}, we provided a larger list of existing metrics for participants to choose from (8 metrics covering group and individual fairness), compared to most user studies \cite{2024crowdsource/10.1145/3640543.3645209, 10.1145/3411764.3445308, Saxena/10.1145/3306618.3314248}. Yet, participants still expressed a clear interest in designing custom metrics tailored to their needs. Our results show that participants often combined multiple existing metrics to construct their own fairness notions, which also reflects the need for more fine-grained and comprehensive fairness assessments discussed above. This an area that has received relatively little attention so far. While previous work has speculated on the need for expressing these custom fairness metrics \cite{doi:10.1080/10447318.2022.2067936}, our work is the first evidence that this is required by stakeholders. This finding may help guide future work by supporting stakeholders to express custom fairness metrics, beyond simply choosing one metric from a set of predefined options. 

At the same time, there is a need for AI experts to explore ways to handle combinations of fairness metrics, as no unified guidance exists for metric selection, and a single metric rarely captures the complexity of real-world fairness needs \cite{10.1145/3457607}. From this perspective, a human-centered approach to fairness may offer a promising direction.

However, such metric combinations create operational challenges for AI experts, as many fairness metrics are mathematically incompatible and optimizing one often degrades another \cite{10.1145/3457607, conflict/10.1145/3351095.3372864, 10.1145/2090236.2090255, 10.1145/3494672}. In practice, such custom ``combinations'' of fairness metrics could be implemented in several ways: through a logical operator `AND' (all selected metrics must be satisfied), a weighted multi-objective approach (metrics contribute with assigned weights) \cite{10.1145/3375627.3375862}, or a constraint hierarchy (higher-priority metrics take precedence) \cite{earnfairness}. When tensions arise among the selected metrics, the trade-off could be visualized, such as through a Pareto frontier, helping AI experts or stakeholders navigate the space of workable compromises in deployment \cite{10.1145/3357236.3395528}.

\subsubsection{Tailoring metrics to features} Unlike the ``one-size-fits-all'' approach commonly adopted by AI experts \cite{de2025towards, hort2023bias}, we observed that a relatively small number of stakeholders used the same metric across all features, while most applied different fairness metrics depending on the specific feature under consideration. Notably, this pattern has not clearly emerged in prior research \cite{10.1145/3411764.3445308}, which often offers limited metric choices or discusses metric preferences in isolation from specific features. In child maltreatment risk prediction, where, within a relatively smaller set of protected features and fairness metrics, participants tended to maintain a consistent metric preference across features \cite{10.1145/3411764.3445308}. This difference further reinforces the need for fairness-assessment tools that not only support context-specific reasoning but also surface feature-specific fairness needs.

\subsubsection{Threshold setting as a reflection of high fairness expectations} Considering that fairness metrics differ not only across categories (e.g., group vs. individual fairness) but also within them (e.g., Counterfactual Fairness vs. Consistency) \cite{10.1145/3457607}, we carefully defined thresholds as the maximum allowable deviation from the optimal fairness outcome of the current metric, for participants, they were maximum unfairness participants were willing to tolerate (0–100\%) to reduce cognitive load while keeping the concept clear and consistent. Although no unanimous agreement on specific fairness threshold values emerged due to personal preferences, participants consistently showed strict settings, revealing high fairness expectations and low tolerance for unfairness. 

We also found that some participants initially aspired to `perfect' fairness outcomes for their chosen feature–metric pairs but introduced tolerance when considering practical feasibility. Importantly, tolerance for unfairness decreased as their perceived harm increased. This pattern echoes prior research \cite{earnfairness} and therefore suggests that stakeholders may set thresholds by balancing high fairness expectations with practical feasibility. Despite this, AI expert practice typically allows for larger deviations (10\% or 20\%), often due to legal, operational, or organizational constraints \cite{10.1145/2783258.2783311, Gupta2021TransitioningFR}, than the thresholds preferred by participants (mostly below 6\%). This gap suggests that AI experts may face greater practical demands during bias mitigation when attempting to align their models with stakeholder expectations.

However, current threshold definitions (i.e., the maximum allowable deviation from a metric's ideal fairness outcome, expressed as 0–100\%) might lack comparability across heterogeneous fairness metrics. Normalization schemes could help when a metric's value range is clearly defined. For instance, normalizing all fairness metrics to a common scale (e.g., 0-1) would make their threshold values directly comparable and help AI experts to better analyze which metrics, or which properties of those metrics, influence stakeholders to set more stringent or more permissive fairness thresholds. \rev{Even after normalization, fairness metrics may differ in whether larger (i.e., close to 1) or smaller values (i.e., close to 0) indicate greater fairness. To make thresholds comparable across all metrics, we could adopt a single convention in which lower values always represent better fairness and values below a chosen threshold are considered acceptable. Metrics that originally follow the opposite direction are transformed by reversing their scale: a normalized value \(x\) used for thresholding is replaced with \(1 - x\) so that, for example, a threshold value originally considered fair at \(\geq 0.9\) becomes an equivalent fair value at \(\leq 0.1\).} However, a metric's value range is not always straightforward to determine, especially when it is unbounded and lacks a known maximum, for example, in the case of stakeholder-defined custom fairness metrics. This can limit the applicability of many normalization schemes and introduce additional effort for AI experts.

\subsubsection{Fairness priorities are not always static.} We observed shifts in participants' fairness priorities as they engaged in assessing fairness. Such dynamics may be driven by considerations of contextual relevance and trade-offs as participants are exposed to more fairness- and context-related information. This dynamism also resonates with recent calls from AI experts, realizing that fairness assessment is not a static setting \cite{10.1145/3597199}, though it remains underexplored in stakeholder fairness assessments. This warrants further investigation and should be carefully considered in the design of long-term fairness assessment tools.

\subsection{Design Implications for Tools Supporting and Scaling Stakeholder Fairness Assessments}
\label{sec: functionality and improving usability}
In real-world practice, stakeholder involvement often requires engaging a significantly larger and more diverse population than was feasible in our study. Our current approach, where researchers interact with participants one-on-one in person, is neither scalable nor efficient due to substantial time and resource demands. Consequently, real-world deployment calls for tools that are accessible online and support independent use by multiple stakeholders. However, the latter remains challenging and underexplored \cite{10.1145/3689904.3694698, earnfairness, doi:10.1080/10447318.2022.2067936, tahaei2023systematic}. 

Recent work highlights the need for fairness assessment tools that help stakeholders move from ``aspirational fairness'' (``what (fairness) should be'') to a more technical ``measurement view'' (``how is it working or performing'') \cite{10.1145/3757647}, and that present fairness concepts to the general public with minimal prior knowledge \cite{tahaei2023systematic}. Building on these insights, we suggest that on one hand, tools should support stakeholders in assessing AI fairness in ways that accurately reflect their needs (functional); on the other hand, tools should also communicate fairness concepts clearly and support independent use (informational), thereby enabling scaling.

We outline the first iteration of design implications in Table~\ref{tab:UI design implications} based on our prototype that speaks to both functional and informational needs, supporting and scaling lay stakeholders in assessing AI outcome fairness. These implications are grounded in the stakeholder assessment patterns we observed (Section~\ref{sec: patterns}) and the challenges participants reported during the study (Section~\ref{sec: experience}).

\begin{table*}[ht]
  \caption{Design Implications for Tools Supporting and Scaling Lay Stakeholder Fairness Assessments}
  
  \label{tab:UI design implications}
  \resizebox{\textwidth}{!}{%
    
    \begin{tabular}{l|l|p{4cm}|p{7cm}}
    \hline
    \textbf{Category} & \textbf{Use} & \textbf{Design Implications} & \textbf{Description}\\
    \hline
    
    \multirow[c]{5}{*}{\makecell[l]{\textbf{Feature Selection} \\ (Whom to Protect)}}
    & Functional & Support diverse feature selection & Allow stakeholders to add or remove both protected and non-protected features, rather than defaulting to protected features only. \\
    & & Enable feature ranking & Allow stakeholders to rank features by perceived priority. \\
    \cline{2-4}
    & Informational &  Provide clear and accessible feature descriptions &  Provide descriptions for all features to help stakeholders, including those unfamiliar with the application domain, clearly understand each feature.\\
    &  & Support navigation within large feature sets &  For high-dimensional datasets, avoid unstructured feature lists, for example by presenting related features in coherent orders, and surface manageable subsets when needed while maintaining scenario completeness and avoiding expert-driven selection bias \cite{10.1145/3411764.3445308, 10.5555/3504035.3504042, 2023/10.1145/3579601}.\\
    &  & Scaffold user reflection on fairness &  Help elicit fairness concerns for those unsure where to begin reasoning about fairness without leading them, for example through reflection on experiences of unfairness \cite{10.1145/3630106.3659044}, or historical bias.\\
    \hline
    
    \multirow[c]{7}{*}{\makecell[l]{\textbf{Metric Selection} \\ (How to Protect)}}
    & Functional & Offer diverse metrics & Support both group- and individual-level fairness metrics, including both widely used and less common ones, to avoid defaulting only to expert preferences. \\
    & & Support custom metrics & Allow stakeholders to combine multiple metrics, specify their logical relationships (e.g., and/or), assign weights, indicate metric priorities, or provide free-text input to flexibly express desired fairness metric. \\
    & & Enable feature-metric mapping & Allow linking metrics to features, with flexibility to revise. \\
    & & Intuitive metric explanations & Provide intuitive, graphics-first metric explanations that avoid text-heavy or overloaded visualizations. \\
    & & Allow dynamic assessment & Allow continuous modification of features, metrics, and rankings during the fairness assessment process. \\
    \cline{2-4}
    & Informational & Offer multi-modal explanations while minimizing information load. &  Offer multi-modal explanations for each fairness metric, such as lay-language descriptions, visual examples of when a metric is satisfied or violated, result visualizations, etc., and surface them progressively to reduce information load and address the challenge of knowing where to focus.\\
    & & Offer a comparative view of multiple fairness metrics &  Highlight key characteristics of multiple metrics within the metric comparison view, for example their strengths, limitations, and distinctions, allowing stakeholders to compare them easily without relying on memory or switching between metrics.\\

    \hline
    \multirow[c]{3}{*}{\makecell[l]{\textbf{Threshold Setting} \\ (To What Extent)}}
    & Functional & Thresholds per feature-metric pair & Allow threshold input as either a single value or a range. \\
    \cline{2-4}
    & Informational & Explain fairness thresholds & Describe what the fairness threshold represents and show its valid range for the selected feature–metric pair to support stakeholders' understanding. \\
    &  & Provide real-time fairness assessment feedback & Show instantly whether the model is fair under the current threshold to validate stakeholders' understanding and ensure the threshold aligns with their intent \cite{earnfairness}. \\
    \hline
   
    \end{tabular}
     }
\end{table*}

\subsection{Input to Guide AI Experts}
The diversity and complexity of our stakeholders' fairness decision-making present a significant challenge in determining which choices should ultimately guide an AI expert in mitigating fairness. Therefore, it is necessary to consider multiple possible approaches to effectively manage these complexities arising from different stakeholders. 

One approach is to simplify the complexity across many stakeholders, either through technical or social means. Technically, it would be possible to select the most popular feature-metric pairings to be assessed and mitigated, with thresholds set as the mode value. Of course, this might not guard against unfairness if biases are systemic within societal values. How to guard against unfairness arising from these societal biases is still an open research question.

Selection of features and metrics could also be handled through a social process of achieving collective consensus, as in the EARN Fairness Framework \cite{earnfairness} where stakeholders can engage in negotiations to agree on features and metrics. However, this approach involving stakeholders in direct negotiations does not scale up well, requires careful management of the risk of ``negotiation deadlock'', and still does not alleviate systemic biases. 

Another approach is to embrace this complexity and try to satisfy \textit{all} stakeholders' fairness demands. For example, some approaches \cite{Yokota2022Toward,taka2024human} use ``crowdsourcing'' to elicit stakeholder preferences, which could then be combined using multi-objective optimization \cite{10.1145/3564285} or simply be integrated by using feedback as training examples \cite{taka2024human}. However, these approaches are in their infancy and warrant more research attention.

\section{Conclusion}
We investigated the complexity of stakeholders' fairness assessments for outcome fairness by examining not only \textit{who they want to protect} (feature selection and prioritization), but also \textit{how they want to protect them} (feature–metric pairing), and \textit{what level of unfairness} they consider acceptable (threshold setting). To support this process, we developed a prototype system and conducted a qualitative study in a credit rating scenario with 26 participants. Our findings show that lay stakeholders can express fairness assessments, and do so in complex ways with fairness considerations that are contextually grounded, broad in scope, dynamic, customized, and held to high expectations: 
\begin{itemize}
    \item Participants considered a broad range of features, including protected/sensitive as well as unprotected features. 
    \item Participants were able to prioritize these features but prioritizations were not static; stakeholders might change how they rank features as they engage more deeply in fairness assessments.
    \item Participants tailored metrics for each feature rather than relying on a ``one-size-fits-all'' metric selection approach.
    \item Participants wanted to customize metrics to express complex fairness demands. In some cases this meant meeting two metrics at the same time, in others they wanted to combine aspects of different metrics to create a new metric.
    \item Participants tended to set fairness thresholds that were strict yet varied, reflecting high expectations for AI fairness while also being influenced by personal preferences. 
\end{itemize}

These findings emphasize the importance of a human-centered approach to AI fairness, which accommodates diverse stakeholders without AI expertise. Our results indicate that this is not a trivial pursuit, as stakeholders' perspectives and decision-making processes are dynamic and complex. Therefore, we need new approaches and tools to help align technical fairness with stakeholder and broader social values. Our work sets a direction for achieving this in practical settings.

\begin{acks}
  We gratefully acknowledge the funding provided by the University of Glasgow and Fujitsu Limited. This work was also partially supported by the Engineering and Physical Sciences Research Council [grant number EP/Y009800/1], through funding from Responsible Ai UK (KP0011). We thank all participants for their contributions.
\end{acks}

\bibliographystyle{ACM-Reference-Format}
\bibliography{references}

\clearpage
\onecolumn
\appendix

\section{Appendix}
\subsection{German Credit Dataset: Features and Feature Descriptions}
\label{app:german_credit_features}
Table~\ref{tab:german_features} shows 20 features with their descriptions from the German Credit Dataset that participants could refer to.

\begin{table*}[ht]
\caption{German Credit Dataset: Features with their Descriptions}
\label{tab:german_features}
\centering
\resizebox{\textwidth}{!}{\begin{tabular}{llll}
\hline

Feature Name & Feature Description & Type & Example Value \\ \hline

Duration & Duration of the credit in months & Numerical & 30 (Months) \\
Credit Amount & Credit amount asked by the applicant & Numerical & 5234 (DM) \\
Installment Rate & Installment rate in percentage of disposable income & Numerical & 4 \\
Residence Length & Length of time (in years) the applicant has lived in the present residence & Numerical & 1 \\
Existing Credits & Number of existing credits at this bank & Numerical & 1 \\
Dependents & Number of people being liable to provide maintenance for & Numerical & 1 \\
Age & The age of the applicant in years & Numerical & 28 \\
Gender & The gender of the applicant & Categorical & Male \\
Checking Account & Status/balance of checking account at this bank & Categorical & 0-200 DM \\
Credit History & Past credit history of applicant at this bank & Categorical & Other Existing Credits \\
Purpose & The applicant's purpose for the loan (e.g., car, education) & Categorical & Car (New) \\
Savings & Savings accounts/bonds at this bank & Categorical & <100 DM \\
Employment & Present employment since & Categorical & Unemployed \\
Debtors & Other debtors/guarantors present & Categorical & None \\
Property & Properties that applicant has & Categorical & Car/Other \\
Installment Plans & Other installment plans the applicant is paying & Categorical & None \\
Housing & Housing (rent, own, for free) & Categorical & Own \\
Job & Current job information (unemployed, (un)skilled, management) & Categorical & Management/Officer \\
Telephone & Is there any telephone registered for this applicant? & Binary & Yes \\
Foreign Worker & Is applicant foreign worker? & Binary & Yes \\

\hline

\end{tabular}}
\end{table*}
\clearpage

\subsection{Participant Information}
\label{app:Participant Information}

\begin{table}[ht]
\caption{Participant Information: Participants are grouped by fairness perspective (outcome fairness vs. other perspectives) rather than listed by PID to support easier cross-referencing with results; PID 9, 22, and 25 were also placed in a separate set, since part of their data did not reflect the outcome fairness perspective (Section~\ref{sec: different perspectives}).}
\vspace{-1em}
\label{tab:Participant}
\centering
\resizebox{\textwidth}{!}{\begin{tabular}{llllllll}
\hline
\multicolumn{7}{c}{\textbf{Participants with Outcome Fairness Perspective}}\\
\hline
PID & Loan Application Experience & Age & Gender & Education & Profession & Nationality \\
\hline
1 & No & 35 & Male & Doctoral Degree & Biology & Saudi Arabian \\
2 & No & 23 & Female & Bachelor's Degree & Healthcare and Medical & Nigerian \\
3 & No & 23 & Male & Bachelor's Degree & Construction and Manufacturing & Chinese \\
4 & Yes & 24 & Female & Bachelor's Degree & Computer and IT & Indian \\
6 & Yes & 29 & Male & Master's Degree & Computer and IT & Indian \\
7 & Yes & 24 & Female & Bachelor's Degree & Healthcare and Medical & Iranian \\
10 & Yes & 27 & Male & Bachelor's Degree & Engineering & Indian \\
11 & No & 24 & Female & Bachelor's Degree & Media and Communications & Chinese \\
12 & No & 24 & Male & Bachelor's Degree & Computer and IT & Chinese \\
15 & Yes & 46 & Female & Bachelor's Degree & Healthcare and Medical & Indian \\
16 & No & 21 & Female & Secondary school diploma or equivalent & Business and Finance & Indian \\
17 & No & 26 & Female & Bachelor's Degree & Education & Egyptian \\
18 & Yes & 27 & Male & Master's Degree & Engineering & Indian \\
19 & No & 28 & Female & Master's Degree & Business and Finance & Chinese \\
20 & Yes  & 28 & Female & Master's Degree & Healthcare and Medical & Irish \\
21 & Yes & 76 & Male & Bachelor's Degree & Business and Finance & British \\
23 & No & 28 & Female & Master's Degree & Healthcare and Medical & Chinese \\
24 & No & 23 & Male & Bachelor's Degree & Business and Finance & Chinese \\
26 & Yes & 37 & Male & Bachelor's Degree & Education & Iranian \\
27 & No & 23 & Female & Bachelor's Degree & Physics & British \\
28 & No & 22 & Female & Bachelor's Degree & Education & Chinese \\
29 & No  & 25 & Female & Bachelor's Degree & Healthcare and Medical & Bangladeshi \\
30 & No & 33 & Male & Bachelor's Degree & Education & Bangladeshi \\
\multicolumn{7}{l}{\dotfill} \\
9 & No & 34 & Female & Doctoral Degree & Education & Indian \\
22 & Yes & 35 & Male & Master's Degree & Education & British \\
25 & Yes & 37 & Female & Bachelor's Degree & Chemical Biology & Iranian \\
\hline
\multicolumn{7}{c}{\textbf{Participants with Other Fairness Perspectives}}\\
\hline
PID & Loan Application Experience & Age & Gender & Education & Profession & Nationality \\
\hline
5 & No & 43 & Male & Bachelor's Degree & Political and Social Studies & Pakistani \\
8 & Yes & 33 & Female & Doctoral Degree & Engineering & Chinese \\
13 & Yes & 29 & Female & Bachelor's Degree & Healthcare and Medical & Chinese \\
14 & Yes & 32 & Male & Bachelor's Degree & Business and Finance & Indian \\
\hline
\end{tabular}}
\end{table}

\begin{table}[ht]
\caption{Participant Information Summary: Participants with outcome fairness perspective were included in the analysis, whereas others were excluded.}
\vspace{-1em}

\label{tab: Participant include and exclude}
\centering
\resizebox{\textwidth}{!}
{\begin{tabular}{llllll}
\hline
\multicolumn{6}{c}{\textbf{Summary of Participants with Outcome Fairness Perspective (N=26)}} \\
\hline
 Loan Application Experience & Age & Gender & Education & Profession & Nationality \\
\hline
 Yes (11) & aged 20–30 years (18) & Male (11) & Secondary school diploma or equivalent (1) & Biology (1) & Saudi Arabian (1) \\
  No (15) & aged 30–40 years (6) & Female (15) & Bachelor's Degree (17) & Healthcare and Medical (6) & Nigerian (1) \\
    & aged 40–50 years (1) &  & Master's Degree (6) & Construction and Manufacturing (1) &  Chinese (7) \\
    & over 50 years (1) &  & Doctoral Degree (2) & Computer and IT (3) & Indian (7) \\
    &  &  &  & Education (6) & Iranian (3) \\
    &  &  &  & Engineering (2) & Egyptian (1) \\
    &  &  &  & Business and Finance (4) & Irish (1) \\
    &  &  &  & Chemical Biology (1) & British (3) \\
    &  &  &  & Physics (1) & Bangladeshi (2) \\

\hline
\multicolumn{6}{c}{\textbf{Summary of Participants with Other Fairness Perspective (N=4)}} \\
\hline
 Loan Application Experience & Age & Gender & Education & Profession & Nationality \\
\hline
 Yes (3) & aged 20–30 years (1) & Male (2) & Bachelor's Degree (3) & Political and Social Studies (1) & Pakistani (1) \\
  No (1) & aged 30–40 years (2) & Female (2) & Doctoral Degree (1) & Engineering (1) & Chinese (2) \\
    & aged 40–50 years (1) & & & Healthcare and Medical (1) & Indian (1)\\
      & & & & Business and Finance (1) & \\
\hline
\end{tabular}}
\end{table}
\clearpage

\subsection{Post Questionnaires}
\label{app:Post Questionnaires}

\textbf{Part 1: System Usability Scale (Please rate from 1-5 for each)}

\textit{When you are using this system for seeking ideal fairness metrics and conducting a fairness audit:}

\begin{enumerate}[label=\arabic*.]
    \item[1.] I think that I would like to use this system frequently.
    
    1 Strongly disagree \hspace{0.7cm} 2 Disagree \hspace{0.7cm} 3 Neutral \hspace{0.7cm} 4 Agree\hspace{0.7cm} 5 Strongly Agree

    \item[2.] I found the system unnecessarily complex.   
    
    1 Strongly disagree \hspace{0.7cm} 2 Disagree \hspace{0.7cm} 3 Neutral \hspace{0.7cm} 4 Agree\hspace{0.7cm} 5 Strongly Agree

    \item[3.] I thought the system was easy to use.
    
    1 Strongly disagree \hspace{0.7cm} 2 Disagree \hspace{0.7cm} 3 Neutral \hspace{0.7cm} 4 Agree\hspace{0.7cm} 5 Strongly Agree

    \item[4.] I think that I would need the support of a technical person to be able to use this system.
    
    1 Strongly disagree \hspace{0.7cm} 2 Disagree \hspace{0.7cm} 3 Neutral \hspace{0.7cm} 4 Agree\hspace{0.7cm} 5 Strongly Agree

    \item[5.] I found the various functions in this system were well integrated.
    
    1 Strongly disagree \hspace{0.7cm} 2 Disagree \hspace{0.7cm} 3 Neutral \hspace{0.7cm} 4 Agree\hspace{0.7cm} 5 Strongly Agree

    \item[6.] I thought there was too much inconsistency in this system.
    
    1 Strongly disagree \hspace{0.7cm} 2 Disagree \hspace{0.7cm} 3 Neutral \hspace{0.7cm} 4 Agree\hspace{0.7cm} 5 Strongly Agree

    \item[7.] I would imagine that most people would learn to use this system very quickly.
    
    1 Strongly disagree \hspace{0.7cm} 2 Disagree \hspace{0.7cm} 3 Neutral \hspace{0.7cm} 4 Agree\hspace{0.7cm} 5 Strongly Agree

    \item[8.] I found the system very cumbersome to use.
    
    1 Strongly disagree \hspace{0.7cm} 2 Disagree \hspace{0.7cm} 3 Neutral \hspace{0.7cm} 4 Agree\hspace{0.7cm} 5 Strongly Agree

    \item[9.] I felt very confident using the system.
    
    1 Strongly disagree \hspace{0.7cm} 2 Disagree \hspace{0.7cm} 3 Neutral \hspace{0.7cm} 4 Agree\hspace{0.7cm} 5 Strongly Agree

    \item[10.] I needed to learn a lot of things before I could get going with this system.
    
    1 Strongly disagree \hspace{0.7cm} 2 Disagree \hspace{0.7cm} 3 Neutral \hspace{0.7cm} 4 Agree\hspace{0.7cm} 5 Strongly Agree

\end{enumerate}
\vspace{1em}
\textbf{Part 2: Open-Ended Questions}

\begin{enumerate}[start=11,label=\arabic*.]
    \item[11.] What difficulties and challenges did you encounter during the process of deciding on what is fair?
    
    \item[12.] What additional features or information do you think are needed to better support you in assessing AI fairness?

    \item[13.] What did you particularly like about the user interface?
    
    \item[14.] What do you think could be improved for the user interface?
    
\end{enumerate}
\clearpage

\subsection{Prototype System - Detailed Description}
\label{app:prototype system design}
\subsubsection{Supplementary Material for Task 1: Feature Selection \& Ranking}
Participants were able to click on ``Causal Graph: Training Data'' to check the training data's casual graph. Each node's (feature) influence on the other nodes was shown by arrows, with the thickness of these arrows indicating the strength of the relationship. Orange nodes indicated protected features, whereas the dark blue node represented the applicant's Rated Credit (Ground truth label). Participants were able to click on each node to highlight the influence other nodes had on the selected node and the influence the selected node had on the other nodes via the direction of the arrows. Similarly, Participants could click on ``Causal Graph: Test Data'' and interacted with the test data causal graph with the dark blue node indicating AI's Predicted Credit. In addition to exploring data bias, participants could click on ``Feature Importance'' to examine potential model biases. Our prototype displayed a bar chart with feature importance weights on the X-axis and features on the Y-axis, sorted by importance. Green bars (weights > 0) indicated positive impacts, red bars (weights < 0) negative impacts. Participants could also click on ``Data Distribution'' to view multiple pie charts, each representing the distribution of a feature in the training data.

\begin{figure}[h]
\centering
\includegraphics[width=0.85\textwidth]{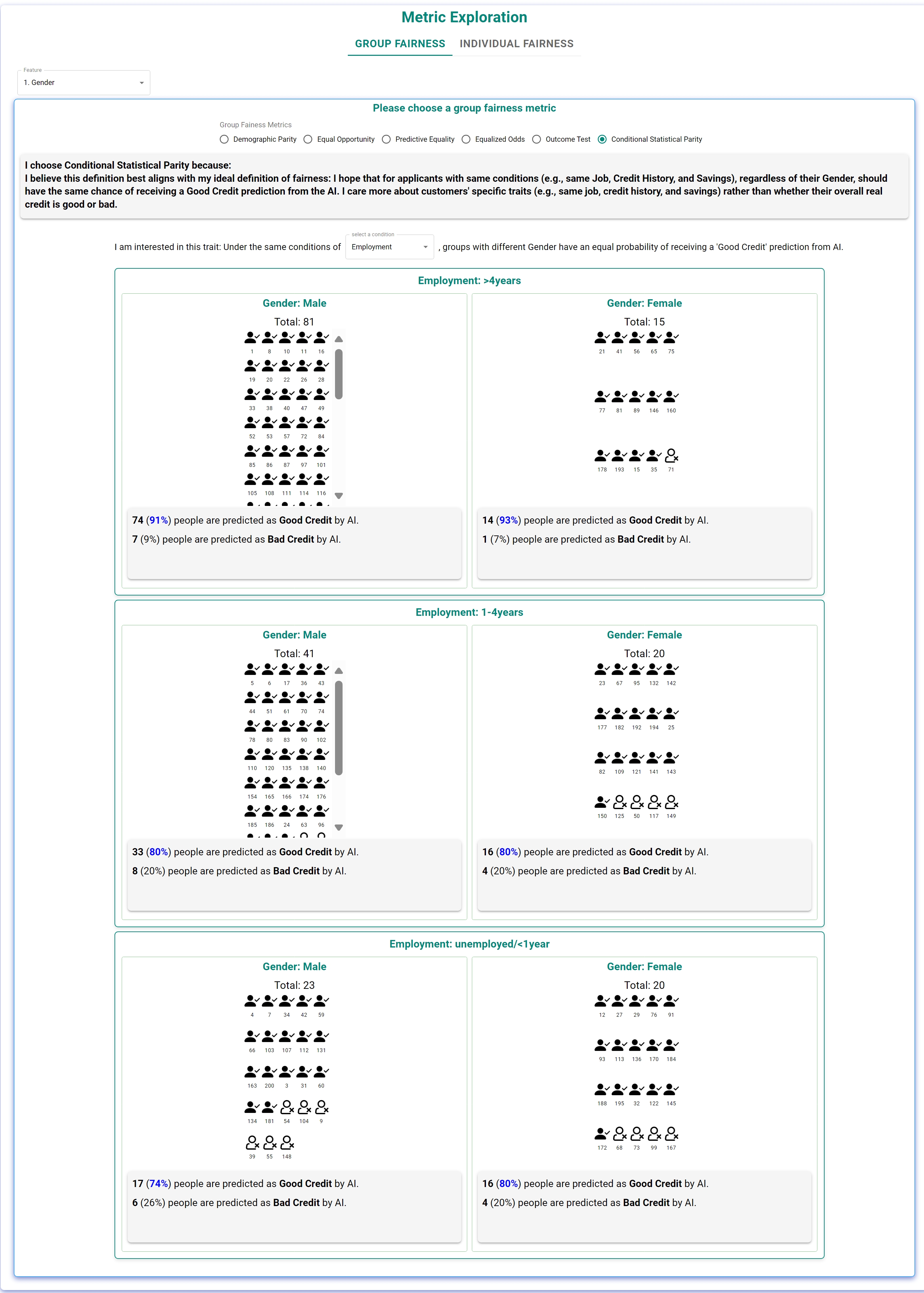}
\caption{Conditional Statistical Parity Metric Explanation}
\label{fig: CSP_metric_info}
\Description{Explanation of the Conditional Statistical Parity metric, combining textual description with an intuitive instance-level visualization.}
\end{figure}

\subsubsection{Supplementary Material for Task 2: Mapping Features to Metrics with Thresholds}
Within the ``Metric Exploration'' component, participants were able to start by clicking “GROUP FAIRNESS” or “INDIVIDUAL FAIRNESS” to choose the fairness category to explore. When exploring group fairness, participants could select a feature from the dropdown menu, arranged by their own feature ranking. Once selected, participants could view the performance of the chosen feature in terms of group fairness. Participants could click on a specific metric, the explanation content would be updated correspondingly, displaying a plain-language description. After clicking ``Conditional Statistical Parity'', participants could customize the condition feature by clicking the dropdown menu, then view results and instance-level visualizations based on the selected condition and feature (Figure ~\ref{fig: CSP_metric_info}).

\begin{figure}[h]
\centering
\includegraphics[width=0.85\textwidth]{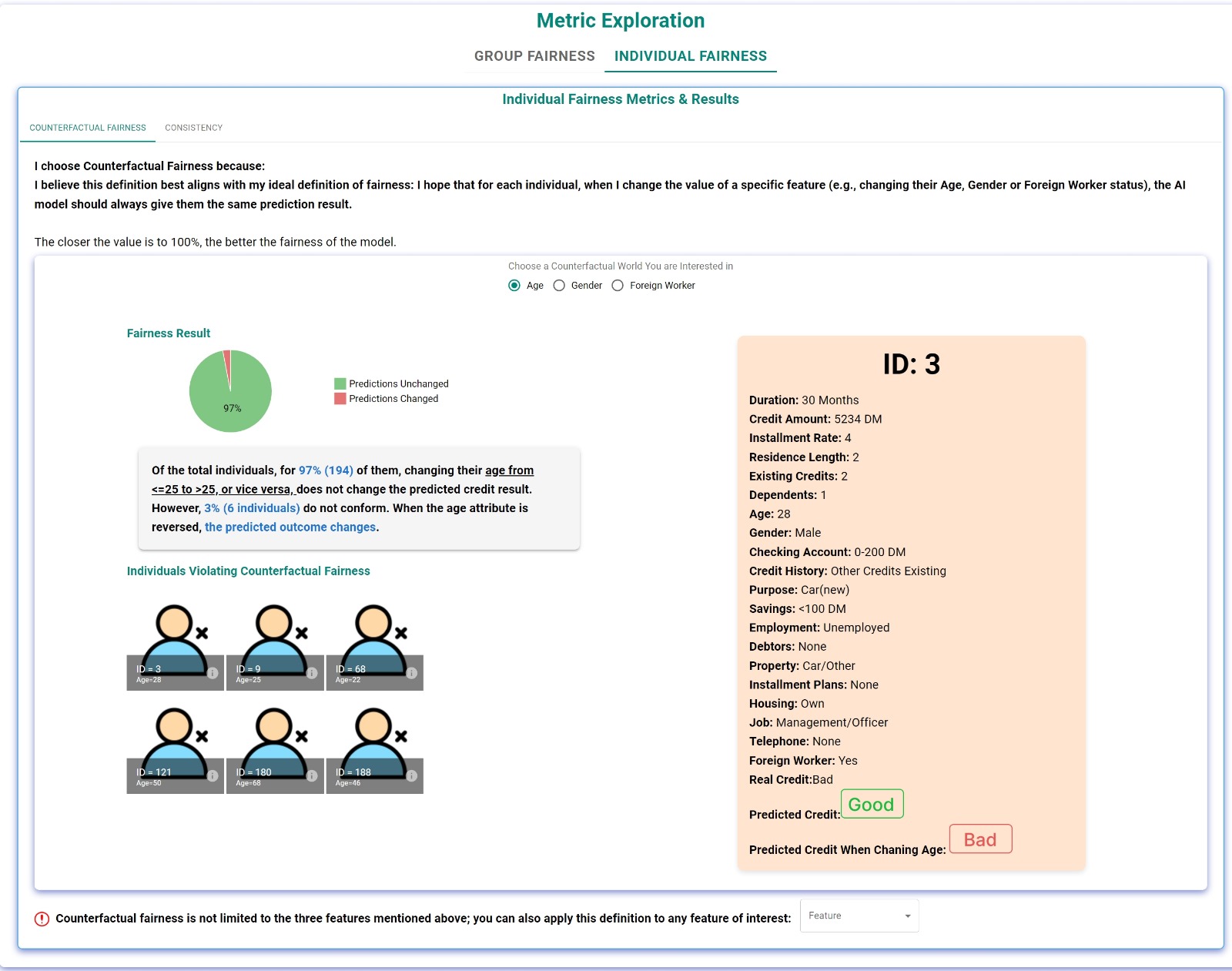}
\caption{Counterfactual Fairness Metric Explanation}
\label{fig: CounterfactualFairness_info}
\Description{Explanation of the Counterfactual Fairness metric, combining textual description with a supporting visualization to illustrate how outcomes differ under counterfactual conditions.}
\end{figure}

When exploring individual fairness, participants were able to click on ``COUNTERFACTUAL FAIRNESS'' and chose protected and sensitive features via radio buttons (Figure ~\ref{fig: CounterfactualFairness_info}). This displayed visualizations and explanations of counterfactual fairness, highlighting examples of individuals who violate counterfactual fairness. Participants could also select any non-protected feature using the dropdown in our prototype system, indicating their interest in the corresponding counterfactual fairness for that feature. For consistency, following FEE's design, participants could click “CONSISTENCY” to see metric results and the predictions of each instance's neighbors, with red dots indicating good predictions and green for the opposite.

\subsubsection{Supplementary Material for Task 3: Fairness Re-Ranking}
\begin{figure}[h]
\centering
\includegraphics[width=0.85\textwidth]{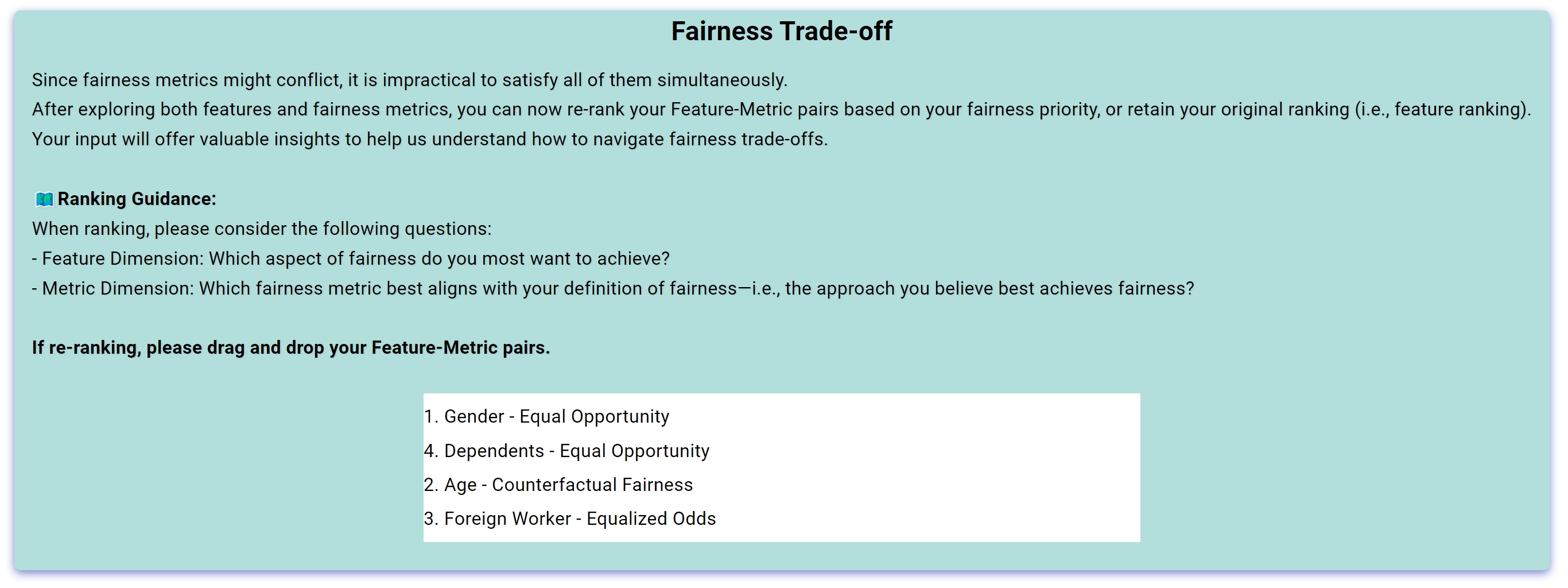}
\caption{Prototype System: Fairness Trade-Off Module (Task 3)}
\label{fig: TradeOff_Reranking_22}
\Description{Prototype system module for Task 3. The top section presents a textual task description, and the bottom section lists feature–metric pairs that participants can rank to explore fairness trade-offs.}
\end{figure}

Figure~\ref{fig: TradeOff_Reranking_22} was designed to help participants navigate fairness trade-offs after exploring all features and fairness metrics. Participants could check their selected Feature-Metric pairs in the white box, ordered by feature ranking and accompanied by numerical labels. Participants could refer to the \textit{Ranking Guidance} and consider two dimensions: from the \textit{Feature Dimension}, which aspect of fairness they most wanted to achieve, and from the \textit{Metric Dimension}, which fairness metric best aligned with their definition of fairness.  

\clearpage

\subsection{Fairness Metrics and Explanations}
\label{app:Fairness Metrics and Explanations}
As referenced in \cite{earnfairness}, Table~\ref{tab:Metrics} provides definitions and calculations for all eight metrics used in our study.

\begin{table*}[ht]
  \caption{Fairness Metrics and Explanations}
  \label{tab:Metrics}
  \captionsetup{skip=0pt}  
  \small
\resizebox{0.9\textwidth}{!}{%
  \begin{tabular}{|p{0.07\textwidth}|p{0.1\textwidth}|p{0.04\textwidth}|p{0.27\textwidth}|p{0.52\textwidth}|}
    \hline
    Category&Metric&Paper&Description &Calculation \\
    \hline
    \multirow{6}{=}{\parbox{0.07\columnwidth}{Group Fairness\textsuperscript{1}}} & Demographic Parity & \cite{NIPS2017_a486cd07}, \cite{10.1145/2090236.2090255} & Both protected and unprotected group members have an equal likelihood of being classified in the positive category. & \text{Difference} = $P(\hat{Y} = \text{1} \mid \text{G = 0}) - P(\hat{Y} = \text{1} \mid \text{G = 1})$ \\
    \cline{2-5}
    & Equal Opportunity & \cite{10.5555/3157382.3157469} & Both protected and unprotected groups have the same probability that a subject in the positive class is assigned a positive predictive value. & \text{Difference} = $P(\hat{Y} = \text{1} \mid \text{Y = 1, G = 0}) - P(\hat{Y} = \text{1} \mid \text{Y = 1, G = 1})$ \\
    \cline{2-5}
    & Predictive Equality & \cite{10.1145/3097983.3098095} & Both protected and unprotected groups have the same probability that a subject in the negative class is assigned a positive predictive value & \text{Difference} = $P(\hat{Y} = \text{1} \mid \text{Y = 0, G = 0}) - P(\hat{Y} = \text{1} \mid \text{Y = 0, G = 1})$  \\
    \cline{2-5}
    & Outcome Test & \cite{articleoutcometest} & Both protected and unprotected groups have the same probability that a subject with a positive predictive value genuinely belongs to the positive class. & \text{Difference} = $P(\text{Y = 1} \mid \hat{Y} \text{= 1, G = 0}) - P(\text{Y = 1} \mid \hat{Y} \text{= 1, G = 1})$  \\
    \cline{2-5}
    & Equalized Odds & \cite{10.5555/3157382.3157469} & Simultaneously satisfy both Equal Opportunity and Predictive Equality. & 
    
    \text{Difference1} = $P(\hat{Y} = \text{1} \mid \text{Y = 1, G = 0}) - P(\hat{Y} = \text{1} \mid \text{Y = 1, G = 1})$
    
    \text{Difference2} = $P(\hat{Y} = \text{1} \mid \text{Y = 0, G = 0}) - P(\hat{Y} = \text{1} \mid \text{Y = 0, G = 1})$
    
    \text{Difference} = $\max(\text{Difference1}, \text{Difference2})$ \\
    
    \cline{2-5}
    & Conditional Statistical Parity & \cite{10.1145/3097983.3098095} & Subjects in both protected and unprotected groups are equally likely to be classified into the positive predicted class, taking into account a set of legitimate features L. In our context, to facilitate lay stakeholders' understanding, we adopted four non-protected features (i.e., job, savings, employment, and credit history) and calculated the metric value separately for each. & \text{Difference} =$P(\hat{Y} = \text{1} \mid \text{L = l, G = 0}) - P(\hat{Y} = \text{1} \mid \text{L = l, G = 1})$  \\
    \hline
    \multirow{2}{=}{\parbox{0.07\columnwidth}{Individual Fairness}} & Counterfactual Fairness & \cite{NIPS2017_a486cd07} & An individual receives the same decision in both the actual world and a counterfactual world where the individual belonged to a different demographic group. & 
    \(
CFR = \frac{1}{N} \sum_{(X, A_{\text{true}}) \in \text{dataset}} 1 \{p(X, A_{\text{true}}) = p(X, A_{\text{counterfactual}})\}
\). \textsuperscript{2} 
 \\
 \cline{2-5}
    & Consistency & \cite{pmlr-v28-zemel13} & Similar individuals receive similar outcomes. & \(
\text{Consistency} = 1 - \frac{1}{n} \sum_{i=1}^{n} | {\hat{y_i}} - \frac{1}{n\_\text{neighbors}} \sum_{j \in \mathcal{N}_{n\_\text{neighbors}}(x_i)}^{} \hat{y_j}| \). \textsuperscript{3} \\
 \hline
\end{tabular}
}
\begin{flushleft}
\textsuperscript{1} The group fairness Metric results are represented by the difference between groups. \\
- $\hat{Y}$: AI-predicted credit rating. $\hat{Y} = 1$ denotes a prediction of good credit, while $\hat{Y} = 0$ denotes a prediction of bad credit. \\
- $Y$: Actual credit rating label. $Y = 1$ indicates a good credit rating as per the original data, while $Y = 0$ indicates a bad credit rating.\\
- $G$: Group membership variable. $G = 0$ denotes the protected group, while $G = 1$ denotes the non-protected group. \\
Similarly, the subgroup fairness metric results are represented by the maximum difference between any two subgroups. \\
\textsuperscript{2} Counterfactual fairness is quantified by the proportion of instances where decisions remain consistent between the actual and counterfactual worlds. \\
- \(N\) represents the total number of instances in the dataset. \\
- \(X,A_{\text{true}}\) denotes an original instance \(X\) with the actual protected feature value \(A_{\text{true}}\) \\
- \(X,A_{\text{counterfactual}}\) denotes an original instance \(X\) with the counterfactual protected feature value \(A_{\text{counterfactual}}\) \\
- \(p(X, A_{\text{true}})\) presents the model's prediction for the instance \(X,A_{\text{true}}\) \\
- \(p(X, A_{\text{counterfactual}})\) presents the model's prediction for the instance \(X,A_{\text{counterfactual}}\) \\
- \(1\{condition\}\) is an indicator function, returning 1 if the given condition is true, otherwise returning 0. \\
\textsuperscript{3} The calculation of Consistency score is computed based on the similarity of AI predictions among the five nearest neighbors, as per IBM's AI Fairness 360 \cite{8843908}. \\
\end{flushleft}
\end{table*}
\clearpage

\subsection{Codebook for Feature Selection Reasons}
\label{app:feature_selection_reasons_codebook}
Table~\ref{tab:feature_selection_reasons_codebook} presents the codebook for feature selection reasons.

\begin{table*}[ht]
\caption{\textbf{Codebook for Feature Selection Reasons}}
\label{tab:feature_selection_reasons_codebook}
\centering
\resizebox{\textwidth}{!}{\begin{tabular}{>{\raggedright}p{0.10\textwidth} | p{0.16\textwidth} | p{0.33\textwidth} | p{0.34\textwidth} | p{0.04\textwidth} | p{0.03\textwidth}}
\multicolumn{6}{c}{column \textit{Parts} refers to Total Number of Participants and \textit{Refs} refer to Total Number of References}\\
\hline
Theme & Code & Description & Example & Parts. &  Refs. \\
\hline
Fairness and Bias Concerns 
& Affects people who are having fresh start 
& Describes concerns regarding fairness for individuals such as international students or foreign workers who need to establish themselves from scratch in a new system or society. 
& Maybe foreign worker, like international students and like people, they have like a starting point. 
& 1 
& 2 \\

\cline{2-6}

 & Unawareness shouldn't be a discrimination 
 & Emphasizes that individuals unfamiliar with a particular country's norms or rules due to their background or residency status should not be penalized by AI systems. 
 & Because I live in a different country. So uh, and I don't know anything about the loan system here. So so definitely shouldn't discriminate 
 & 3 
 & 3 \\
 
\cline{2-6}

 & Caring for family cannot be discriminated 
 & Points out the necessity for AI systems to accommodate individuals with family caregiving responsibilities without introducing discriminatory bias. 
 & For the dependents, I also notice from that graph[data distribution], most of them have one dependent and only a few of them have two dependents. So, that means all of the applicants have dependents and sometimes it does not matter. You need to care for one or two dependents because they're your family and maybe that's part of your daily routine and one dependents, two different, doesn't change too much. So, I think, yeah, because everyone has families so we shouldn't use that as a reason to reject them 
 & 1 
 & 1 \\
 
\cline{2-6}

 & Concerns about bias 
 & Highlights participant concerns regarding biases within AI systems that may unfairly influence decisions affecting people's lives. 
 & If a lot of decisions are falling back on how many dependents, if it's on the form, how many dependents do you have, then you're more likely, there will be more weight given to that factor, so they could be denied or accepted and the weight would be too strongly based on dependence, not considering other factors, so I think that's really one that's really important to watch out for, because they're so connected to the other ones 
 & 9 
 & 11 \\
 
\cline{2-6}

 & Doesn't fully capture the scenario 
 & Suggests that some features, such as residence length, may fail to comprehensively represent individuals' unique situations and therefore require reconsideration in fairness assessments. 
 & Residence length cannot approve something. Sometimes people just moved to this new city or country. It cannot help us to fully understand this person's repayment. One-sided. 
 & 7 
 & 11 \\

\cline{2-6}

 & Equality for protected features 
 & Stresses the importance of ensuring fairness and equality in AI systems for protected characteristics, particularly to prevent discrimination based on features like residence. 
 & And then residence length, So it's like foreign workers. which means how long they have been living in a country. So yeah, I feel like they shouldn't judge based on that because that would be discriminating against immigrants. And yeah, it shouldn't matter how long they have been staying at a place as long as they have a stable job and it's a stable income. 
 & 14 
 & 16 \\

\bottomrule
\end{tabular}}
\begin{flushright}
Continued on next page.
\end{flushright}
\end{table*}
\clearpage

\begin{table*}[ht]
\centering
\resizebox{\textwidth}{!}{\begin{tabular}{>{\raggedright}p{0.10\textwidth} | p{0.16\textwidth} | p{0.33\textwidth} | p{0.34\textwidth} | p{0.04\textwidth} | p{0.03\textwidth}}
\multicolumn{6}{c}{\textbf{Table 7 continued from previous page}}\\
\hline
Theme & Code & Description & Example & Parts. &  Refs. \\
\hline

 & Discrimination against Inherent Human Traits and Social Labels 
 & Raises issues regarding how factors like residence length might be improperly used to make assessments, potentially creating unfair outcomes for certain groups. 
 & I think the first [selected] ones [Age and Gender] are all inherent human traits. Um. Although the second one [foreign worker] doesn't fit perfectly, they all feel like labels that people naturally carry. If people judge others based on these, it feels a bit like discrimination, like looking at others through tinted glasses. So I rank them higher. 
 & 12 
 & 17 \\
 \cline{2-6}
 & Influenced by other features 
 & Points out the potential for certain features to be misinterpreted by AI systems, leading to unfair outcomes or decisions. 
 & If I have some dependent, I might not own that much, then that will affect my savings. So dependence and savings go together. 
 & 2 
 & 2 \\
 
\cline{2-6}
 
 & Intention matters 
 & Highlights concerns about the potential for AI-driven decisions to inadvertently incorporate bias, particularly in contexts such as residency. 
 & The duration is like I think it falls down to the point where whether the intention of the person who is taking this loan is crucial. So whether the duration is longer or shorter should not be discriminated by a AI. 
 & 3 
 & 3 \\
 
\cline{2-6}
 
 & Not relevant to context-specific decision-making 
 & Focuses on the concern that particular features used in AI models might directly lead to discriminatory or unfair outcomes for affected individuals. 
 & We shouldn't really be discriminating based on age, it should be based on whether or not they can pay back the loan [...].
 & 16 
 & 22 \\
 
\cline{2-6}

 \hline
Financial Ability Considerations 
& Paying long term is fine as long as affordable 
& Emphasizes the need to evaluate individuals' long-term payment capacity as long as it remains affordable, ensuring AI fairness in financial scenarios. 
& Duration is again just the time that they take to pay the loan back, so just like installment rate, I think it should be fine as long as they can pay it back. So if it makes practical sense, then I don't think duration should be taken into account as much. 
& 3 
& 4 \\
 
\cline{2-6}

 & Avoid prejudging people's choices
 & Suggests that personal difficulties and hardships should not be considered negative factors when assessing financial ability. 
 & Purpose I think [...] we cannot prejudge, is that purpose good or bad or is that superior than any other purpose. So, everyone have their own choices. 
 & 8 
 & 8 \\
 
\cline{2-6}
 
 & Stable income is more necessary 
 & Highlights the importance of stable income over other potentially less relevant factors in determining financial capability. 
 & Dependents also, I don't see why it's there. Because, if the person has, okay credit history, the person has a job and can pay and has a salary, has an income, it doesn't matter if he has children, if he has dependents, or not with their children, or parents, or elderly, like, they're taking care of. But, I don't see why they can consider caregivers, for example, not to take loans, or I think it's also a bit discriminatory. 
 & 6 
 & 6 \\
 
\cline{2-6}
 
 & Cannot determine repayment capability 
 & Points out that certain features might not accurately predict repayment ability and could lead to unfair decisions. 
 & Nowadays, people who can earn a lot of money. Part of them are experienced, older business people. Now we have a lot of Internet Celebrity. They are very young. We have many different age ranges. Cannot judge people's credit level on age 
 & 8 
 & 8 \\
 
\cline{2-6}
 
 & Don't capture context-relevant information 
 & Acknowledges the complexity of capturing an individual's complete financial situation through limited features in AI systems. 
 & The whole telephone thing, I think that's just nonsense. Whether someone leaves a phone number or not has just become another data point?
 & 9 
 & 16 \\
 
\bottomrule
\end{tabular}}
\end{table*}

\setcounter{table}{7}

\subsection{Results Table}
\label{app:results table}
The table below presents the fairness decision-making results of 26 participants who held outcome fairness perspective.
\begin{table}[h]
\caption{Fairness decision-making from the outcome fairness perspective by 26 participants, including the final selected feature list (All Features) after re-ranking, paired each Feature with Metric. For features with a ``Custom Fairness'' metric, the Custom Metric Detail row displays the specific metric combinations, and the Threshold represents the fairness threshold corresponding to each feature-metric pair.} 
\label{tab: allFeature_metrics_Threshold_Custom_Fariness}
\centering
\resizebox{\textwidth}{!}{\begin{tabular}{lllllllllllll}
\hline
PID \textsuperscript{1} & Metric Decision& 1st & 2nd & 3rd & 4th & 5th & 6th & 7th & 8th & 9th & 10th & 11th \\
\hline
\multicolumn{13}{c}{\textbf{Outcome Fairness Perspective}}\\
\hline
P1 & All Features & Job & Age & Checking Account & Credit History & Dependents \\
 & Feature & Job & Age & Checking Account & Credit History & Dependents \\
 & Metric & Consistency & Conditional Statistical Parity & Conditional Statistical Parity & Consistency & Counterfactual Fairness \\
 &  Custom Metric Detail &   &   &   &   &   \\
 & Threshold & 10\% & 10\% & 10\% & 10\% & 10\% \\
\hline

P2 & All Features & Gender & Foreign Worker & Telephone \\
 & Feature & Gender & Foreign Worker & Telephone \\
 & Metric & Custom Fairness & Custom Fairness & Counterfactual Fairness \\
 &  Custom Metric Detail & Equal Opportunity \& True Negative Rate & Equal Opportunity \& True Negative Rate&   \\
 & Threshold & 0\% & 0\% & 3\% \\
\hline

P3 & All Features & Gender & Age & Foreign Worker & Telephone & Purpose & Duration \\
 & Feature & Gender & Age & Foreign Worker & Telephone \\
 & Metric & Counterfactual Fairness & Counterfactual Fairness & Counterfactual Fairness & Counterfactual Fairness \\
 &  Custom Metric Detail &   &   &   &   \\
 & Threshold & 5\% & 5\% & 5\% & 5\% &  \% &  \% \\
\hline

P4 & All Features & Foreign Worker & Age & Purpose & Gender & Employment & Duration \\
 & Feature & Foreign Worker & Age & Purpose \\
 & Metric & Conditional Statistical Parity & Predictive Equality & Conditional Statistical Parity \\
 &  Custom Metric Detail &   &   &   \\
 & Threshold & 8\% & 3\% & 20\% &  \% &  \% &  \% \\
\hline

P6 & All Features & Gender & Age & Dependents & Foreign Worker & Telephone & Savings \\
 & Feature & Gender & Age & Dependents & Foreign Worker \\
 & Metric & Custom Fairness & Custom Fairness & Conditional Statistical Parity & Conditional Statistical Parity \\
 &  Custom Metric Detail & Counterfactual Fairness \& Equalized Odds & Counterfactual Fairness \& Outcome Test &   &   \\
 & Threshold & 0\% & 5\% & 5\% & 5\% &  \% &  \% \\
\hline

P7 & All Features & Credit History & Purpose & Foreign Worker & Gender & Age & Telephone \\
 & Feature & Credit History & Purpose & Foreign Worker & Gender \\
 & Metric & Consistency & Custom Fairness & Equal Opportunity & Conditional Statistical Parity \\
 &  Custom Metric Detail &   & Outcome Test \& Counterfactual Fairness &   &   \\
 & Threshold & 40\% & 40\% & 40\% & 40\% &  \% &  \% \\
\hline

P9\textsuperscript{2} & All Features & Age \\
 & Feature & Age \\
 & Metric & Counterfactual Fairness \\
 &  Custom Metric Detail &   \\
 & Threshold & 20\% \\
\hline

P10 & All Features & Telephone & Gender & Savings & Age & Employment & Installment Rate & Existing Credits & Property & Foreign Worker & Debtors & Housing \\
 & Feature & Telephone & Gender & Savings & Age & Employment \\
 & Metric & Custom Fairness & Custom Fairness & Conditional Statistical Parity & Conditional Statistical Parity & Conditional Statistical Parity \\
 &  Custom Metric Detail & Counterfactual Fairness \& Equalized Odds & Counterfactual Fairness \& Equal Opportunity &   &   &   \\
 & Threshold & 0\% & 0\% & 0\% & 15\% & 15\% &  \% &  \% &  \% &  \% &  \% &  \% \\
\hline

P11 & All Features & Telephone & Residence Length & Purpose & Foreign Worker & Dependents & Gender & Job \\
 & Feature & Telephone & Residence Length & Purpose \\
 & Metric & Custom Fairness & Custom Fairness & Custom Fairness \\
 &  Custom Metric Detail & Equal Opportunity \& True Negative Rate & Equal Opportunity \& True Negative Rate & Equal Opportunity \& True Negative Rate \\
 & Threshold & 5\% & 5\% & 5\% &  \% &  \% &  \% &  \% \\
\hline

P12 & All Features & Foreign Worker & Telephone \\
 & Feature & Foreign Worker & Telephone \\
 & Metric & Counterfactual Fairness & Consistency \\
 &  Custom Metric Detail &   &   \\
 & Threshold & 0\% & 3\% \\
\hline

P15 & All Features & Gender & Age & Purpose & Telephone & Residence Length & Foreign Worker \\
 & Feature & Gender & Age & Purpose & Telephone & Residence Length & Foreign Worker \\
 & Metric & Counterfactual Fairness & Counterfactual Fairness & Equalized Odds & Counterfactual Fairness & Counterfactual Fairness & Equalized Odds \\
 &  Custom Metric Detail &   &   &   &   &   &   \\
 & Threshold & 0\% & 0\% & 5\% & 0\% & 0\% & 5\% \\
\hline

P16 & All Features & Telephone & Gender & Job & Installment Rate & Residence Length & Duration & Property & Age \\
 & Feature & Telephone & Gender & Job & Installment Rate & Residence Length & Duration & Property \\
 & Metric & Counterfactual Fairness & Custom Fairness & Counterfactual Fairness & Consistency & Counterfactual Fairness & Consistency & Counterfactual Fairness \\
 &  Custom Metric Detail &   & Consistency \& Counterfactual Fairness &   &   &   &   &   \\
 & Threshold & 10\% & 10\% & 10\% & 10\% & 10\% & 10\% & 10\% &  \% \\
\hline

P17 & All Features & Age & Gender & Foreign Worker & Dependents & Purpose \\
 & Feature & Age & Gender & Foreign Worker & Dependents & Purpose \\
 & Metric & Custom Fairness & Custom Fairness & Custom Fairness & Custom Fairness & Custom Fairness \\
 &  Custom Metric Detail & Counterfactual Fairness \& Equal Opportunity & Counterfactual Fairness \& Equal Opportunity & Counterfactual Fairness \& Equal Opportunity & Counterfactual Fairness \& Equal Opportunity & Counterfactual Fairness \& Equal Opportunity \\
 & Threshold & 10\% & 1\% & 5\% & 5\% & 5\% \\
\hline

P18 & All Features & Gender & Age & Debtors & Employment & Housing & Property \\
 & Feature & Gender & Age & Debtors \\
 & Metric & Custom Fairness & Custom Fairness & Custom Fairness \\
 &  Custom Metric Detail & Consistency \& Counterfactual Fairness 
 & \makecell[l]{
 Consistency \& 
 \\
  Counterfactual Fairness\& 
 \\
  Conditional Statistical Parity }
 & \makecell[l]{
 Consistency \& 
 \\
  Counterfactual Fairness \& 
 \\
  Equalized Odds} \\
 
 & Threshold & 1\% & 5\% & 0\% &  \% &  \% &  \% \\
\hline

P19 & All Features & Gender & Age & Purpose & Job & Telephone & Residence Length & Foreign Worker & Dependents & Housing & Duration \\
 & Feature & Gender & Age & Purpose \\
 & Metric & Predictive Equality & Equal Opportunity & Consistency \\
 &  Custom Metric Detail &   &   &   \\
 & Threshold & 10\% & 15\% & 15\% &  \% &  \% &  \% &  \% &  \% &  \% &  \% \\
\hline

P20 & All Features & Dependents & Gender & Age & Checking Account & Housing & Employment & Job \\
 & Feature & Dependents & Gender & Age & Checking Account & Housing \\
 & Metric & Custom Fairness & Conditional Statistical Parity & Equal Opportunity & Conditional Statistical Parity & Equal Opportunity \\
 &  Custom Metric Detail & Counterfactual Fairness \& Equal Opportunity &   &   &   &   \\
 & Threshold & 1\% & 1\% & 1\% & 5\% & 5\% &  \% &  \% \\
\hline

P21 & All Features & Gender & Telephone & Foreign Worker \\
 & Feature & Gender & Telephone & Foreign Worker \\
 & Metric & Counterfactual Fairness & Counterfactual Fairness & Equalized Odds \\
 &  Custom Metric Detail &   &   &   \\
 & Threshold & 10\% & 10\% & 1\% \\
\hline

P22\textsuperscript{3} & All Features & Gender & Dependents & Age & Foreign Worker \\
 & Feature & Gender & Dependents & Age & Foreign Worker \\
 & Metric & Equal Opportunity & Equal Opportunity & Counterfactual Fairness & Equalized Odds \\
 &  Custom Metric Detail &   &   &   &   \\
 & Threshold & 5\% & 5\% & 3\% & 3\% \\
\hline

P23 & All Features & Gender & Foreign Worker & Age & Dependents & Telephone & Residence Length & Housing & Job \\
 & Feature & Gender & Foreign Worker & Age & Dependents & Telephone & Residence Length & Housing & Job \\
 & Metric & Consistency & Consistency & Consistency & Consistency & Consistency & Consistency & Consistency & Consistency \\
 &  Custom Metric Detail &   &   &   &   &   &   &   &   \\
 & Threshold & 5\% & 5\% & 5\% & 5\% & 5\% & 5\% & 5\% & 5\% \\
\hline

P24 & All Features & Installment Plans & Checking Account & Property & Savings & Job & Credit History & Age & Foreign Worker \\
 & Feature & Installment Plans & Checking Account & Property & Savings & Job & Credit History & Age & Foreign Worker \\
 & Metric & Equal Opportunity & Equal Opportunity & Equal Opportunity & Equal Opportunity & Equal Opportunity & Equal Opportunity & Equal Opportunity & Equal Opportunity \\
 &  Custom Metric Detail &   &   &   &   &   &   &   &   \\
 & Threshold & 20\% & 20\% & 20\% & 20\% & 20\% & 20\% & 20\% & 20\% \\
\hline

P25\textsuperscript{4} & All Features & Telephone & Purpose & Dependents & Foreign Worker & Residence Length & Job \\
 & Feature & Telephone & Purpose & Dependents & Foreign Worker & Residence Length & Job \\
 & Metric & Counterfactual Fairness & Counterfactual Fairness & Conditional Statistical Parity & Counterfactual Fairness & Counterfactual Fairness & Conditional Statistical Parity \\
 &  Custom Metric Detail &   &   &   &   &   &   \\
 & Threshold & 0.5\% & 0.5\% & 0.5\% & 0.5\% & 0.5\% & 0.5\% \\
\hline

P26 & All Features & Gender & Residence Length & Purpose & Foreign Worker & Dependents & Installment Rate & Installment Plans & Duration \\
 & Feature & Gender & Residence Length & Purpose & Foreign Worker & Dependents & Installment Rate & Installment Plans & Duration \\
 & Metric & Counterfactual Fairness & Equalized Odds & Counterfactual Fairness & Equalized Odds & Counterfactual Fairness & Equalized Odds & Equalized Odds & Counterfactual Fairness \\
 &  Custom Metric Detail &   &   &   &   &   &   &   &   \\
 & Threshold & 1\% & 3\% & 2\% & 2\% & 1\% & 3\% & 3\% & 3\% \\
\hline

P27 & All Features & Gender & Age & Foreign Worker & Dependents & Job & Housing & Residence Length & Checking Account \\
 & Feature & Gender & Age & Foreign Worker & Dependents & Job & Housing & Residence Length & Checking Account \\
 & Metric & Custom Fairness & Custom Fairness & Custom Fairness & Custom Fairness & Custom Fairness & Custom Fairness & Custom Fairness & Custom Fairness \\
 &  Custom Metric Detail & \makecell[l]{Consistency\& \\Counterfactual Fairness \& \\Conditional Statistical Parity} & \makecell[l]{Consistency \& \\Counterfactual Fairness \& \\Conditional Statistical Parity} & \makecell[l]{Consistency \& \\Counterfactual Fairness \& \\Conditional Statistical Parity} & \makecell[l]{Consistency \& \\Counterfactual Fairness \& \\Conditional Statistical Parity} & \makecell[l]{Consistency \& \\Counterfactual Fairness \& \\Conditional Statistical Parity} & \makecell[l]{Consistency \& \\Counterfactual Fairness \& \\Conditional Statistical Parity} & \makecell[l]{Consistency \& \\Counterfactual Fairness \& \\Conditional Statistical Parity} & \makecell[l]{Consistency \& \\Conditional Statistical Parity} \\
 & Threshold & 5\% & 5\% & 5\% & 5\% & 5\% & 5\% & 5\% & 20\% \\
\hline

P28 & All Features & Gender & Telephone & Purpose & Age & Dependents & Foreign Worker \\
 & Feature & Gender & Telephone & Purpose & Age & Dependents & Foreign Worker \\
 & Metric & Equalized Odds & Outcome Test & Consistency & Conditional Statistical Parity & Conditional Statistical Parity & Equalized Odds \\
 &  Custom Metric Detail &   &   &   &   &   &   \\
 & Threshold & 1\% & 5\% & 5\% & 5\% & 5\% & 5\% \\
\hline

P29 & All Features & Gender & Foreign Worker & Employment & Age & Residence Length \\
 & Feature & Gender & Foreign Worker & Employment & Age & Residence Length \\
 & Metric & Counterfactual Fairness & Counterfactual Fairness & Conditional Statistical Parity & Consistency & Consistency \\
 &  Custom Metric Detail &   &   &   &   &   \\
 & Threshold & 3\% & 5\% & 10\% & 5\% & 5\% \\
\hline

P30 & All Features & Gender & Foreign Worker & Residence Length & Property & Housing & Age & Telephone & Debtors & Duration \\
 & Feature & Gender & Foreign Worker & Residence Length & Property & Housing & Age & Telephone & Debtors & Duration \\
 & Metric & Equal Opportunity & Conditional Statistical Parity & Equal Opportunity & Counterfactual Fairness & Equal Opportunity & Counterfactual Fairness & Equal Opportunity & Equal Opportunity & Equal Opportunity \\
 &  Custom Metric Detail &   &   &   &   &   &   &   &   &   \\
 & Threshold & 0\% & 5\% & 0\% & 2\% & 0\% & 2\% & 0\% & 0\% & 0\% \\
\hline

\end{tabular}}
\begin{flushleft}
\textsuperscript{1} Participants P5, P8, P13, and P14 held the perspective of affordability fairness. Among them, P5 and P13 highlighted the importance of procedural fairness in pursuing affordability fairness. As noted in Section~\ref{sec:prototype system} and Section~\ref{sec: Results}, our system is designed for exploring fairness and metrics related to outcome fairness. Therefore, the results of these four participants were excluded from the table and our analysis.\\
\textsuperscript{2} P9 mentioned the notion of fairness through unawareness, stating that `Telephone', `Gender', and `Foreign Worker' should be excluded. \\
\textsuperscript{3} P22 mentioned the notion of fairness through unawareness, stating that `Telephone' and `Residence Length' should be excluded. \\
\textsuperscript{4} P25 mentioned the notion of fairness through unawareness, stating that `Installment Plans' should be excluded.
\end{flushleft}
\end{table}

\end{document}